\documentclass{sbc2023}%

\usepackage[caption=false,farskip=0pt]{subfig}
\usepackage{graphicx}
\usepackage[misc,geometry]{ifsym} 

\ifarxiv
    \usepackage{changepage}
\else
    \usepackage{fontspec}
\fi
\usepackage{fontawesome}
\usepackage{academicons}
\usepackage{color}
\usepackage{aas_macros}
\usepackage[bottom]{footmisc}
\usepackage{supertabular}
\usepackage{afterpage}
\usepackage{url}
\usepackage{pifont}
\usepackage{multicol}
\usepackage{multirow}
\usepackage[acronym]{glossaries}
\usepackage[table,dvipsnames]{xcolor}
\usepackage{xspace}
\usepackage[capitalise,noabbrev]{cleveref} %
\usepackage{hyperref}
\setcitestyle{square}
\usepackage{arydshln}
\usepackage{tabularx}
\usepackage{verbatimbox}
\definecolor{orcidlogo}{rgb}{0.37,0.48,0.13}
\definecolor{unilogo}{rgb}{0.16, 0.26, 0.58}
\definecolor{maillogo}{rgb}{0.58, 0.16, 0.26}
\definecolor{darkblue}{rgb}{0.0,0.0,0.0}
\hypersetup{colorlinks,breaklinks,
            linkcolor=darkblue,urlcolor=darkblue,
            anchorcolor=darkblue,citecolor=darkblue}
\usepackage{tabularray}
\usepackage{booktabs}
\usepackage{balance}
\usepackage{siunitx}

\usepackage{ifthen}
\usepackage[normalem]{ulem} %

\newcommand\major[1]{{\textcolor{black}{#1}}}

\usepackage{makecell}   %

\newacronymstyle{long-short-br}
{%
  \GlsUseAcrEntryDispStyle{long-short}%
}%
{%
  \GlsUseAcrStyleDefs{long-short}%
}
\setacronymstyle{long-short-br}

\ifarxiv
\else
    \jid{JBCS}
    \jtitle{Journal of the Brazilian Computer Society, 202X, XX:1, }
    \doi{10.5753/jbcs.202X.XXXXXX}
    \copyrightstatement{This work is licensed under a Creative Commons Attribution 4.0 International License}
    \jyear{202X}
\fi

\title{Toward Advancing License Plate Super-Resolution in Real-World Scenarios: A Dataset and Benchmark}

\ifarxiv
    \newcommand{\authorsAbbrev}{}
\else
    \newcommand{\authorsAbbrev}{Nascimento et al. 2025}
\fi

\ifarxiv

\author[\authorsAbbrev]{
\affil{\textbf{Valfride Nascimento}$^*$~~[~\textbf{Federal University of Paraná}~|\href{mailto:vwnascimento@inf.ufpr.br}{~\textbf{\textit{vwnascimento@inf.ufpr.br}}}~]}

\affil{\textbf{Gabriel E. Lima}~~[~\textbf{Federal University of Paraná~}|\href{mailto:menotti@inf.ufpr.br}~\href{mailto:gelima@inf.ufpr.br}{~\textbf{\textit{gelima@inf.ufpr.br}}}~]}

\affil{\textbf{Rafael O. Ribeiro}~~[~\textbf{Brazilian Federal Police}~|\href{mailto:rafael.ror@pf.gov.br}{~\textbf{\textit{rafael.ror@pf.gov.br}}}~]}

\affil{\textbf{William Robson Schwartz}~~[~\textbf{Federal University of Minas Gerais}~|\href{mailto:william@dcc.ufmg.br}{~\textbf{\textit{william@dcc.ufmg.br}}}~]}

\affil{\textbf{Rayson Laroca}~~[~\textbf{Pontifical Catholic University of Paraná, Federal University of Paraná}~|\href{mailto:rayson@ppgia.pucpr.br}{~\textbf{\textit{rayson@ppgia.pucpr.br}}}~]}

\affil{\textbf{David Menotti}~~[~\textbf{Federal University of Paraná~}|\href{mailto:menotti@inf.ufpr.br}{~\textbf{\textit{menotti@inf.ufpr.br}}}~]}

}

\else

\author[\authorsAbbrev]{
\affil{\textbf{Valfride Nascimento}~\href{https://orcid.org/0000-0002-7416-613X}{\textcolor{orcidlogo}{\aiOrcid}}~\textcolor{blue}{\faEnvelopeO}~~[~\textbf{Federal University of Paraná}~|\href{mailto:vwnascimento@inf.ufpr.br}{~\textbf{\textit{vwnascimento@inf.ufpr.br}}}~]}

\affil{\textbf{Gabriel E. Lima}~\href{https://orcid.org/0009-0009-7599-8550}{\textcolor{orcidlogo}{\aiOrcid}}~~[~\textbf{Federal University of Paraná~}|\href{mailto:menotti@inf.ufpr.br}~\href{mailto:gelima@inf.ufpr.br}{~\textbf{\textit{gelima@inf.ufpr.br}}}~]}

\affil{\textbf{Rafael O. Ribeiro}~\href{https://orcid.org/0000-0002-6381-3469}{\textcolor{orcidlogo}{\aiOrcid}}~~[~\textbf{Brazilian Federal Police}~|\href{mailto:rafael.ror@pf.gov.br}{~\textbf{\textit{rafael.ror@pf.gov.br}}}~]}

\affil{\textbf{William Robson Schwartz}~\href{https://orcid.org/0000-0003-1449-8834}{\textcolor{orcidlogo}{\aiOrcid}}~~[~\textbf{Federal University of Minas Gerais}~|\href{mailto:william@dcc.ufmg.br}{~\textbf{\textit{william@dcc.ufmg.br}}}~]}

\affil{\textbf{Rayson Laroca}~\href{https://orcid.org/0000-0003-1943-2711}{\textcolor{orcidlogo}{\aiOrcid}}~~[~\textbf{Pontifical Catholic University of Paraná, Federal University of Paraná}~|\href{mailto:rayson@ppgia.pucpr.br}{~\textbf{\textit{rayson@ppgia.pucpr.br}}}~]}

\affil{\textbf{David Menotti}~\href{https://orcid.org/0000-0003-2430-2030}{\textcolor{orcidlogo}{\aiOrcid}}~~[~\textbf{Federal University of Paraná~}|\href{mailto:menotti@inf.ufpr.br}{~\textbf{\textit{menotti@inf.ufpr.br}}}~]}

}

\fi

\usepackage{ifthen}
\usepackage{xcolor}

\newboolean{showcomments}
\setboolean{showcomments}{true} %
\newboolean{showcommentsMA}
\setboolean{showcommentsMA}{true} %
\usepackage{footmisc}
\newcommand*{\MyComment}[4][]{
  \ifthenelse{\boolean{showcomments}}{%
    \textcolor{#2}{[\textbf{\ifthenelse{\equal{#1}{}}{#3}{#3(#1)}}: #4]}%
  }{}%
}

\newcommand*{\MyCommentMA}[4][]{
  \ifthenelse{\boolean{showcommentsMA}}{%
    \textcolor{#2}{[\textbf{\ifthenelse{\equal{#1}{}}{#3}{#3(#1)}}: #4]}%
  }{}%
}

\ifthenelse{\boolean{showcomments}}{
  \newcommand*{\DM}[2][]{\MyComment[#1]{orange}{DM}{#2}}
  \newcommand*{\RL}[2][]{\MyComment[#1]{Rhodamine}{RL}{#2}}
  \newcommand*{\GL}[2][]{\MyComment[#1]{brown}{GL}{#2}}
  \newcommand*{\WS}[2][]{\MyComment[#1]{green}{WS}{#2}}
}{%
  \newcommand*{\DM}[2][]{}
  \newcommand*{\RL}[2][]{}
  \newcommand*{\GL}[2][]{}
  \newcommand*{\WS}[2][]{}
}

\ifthenelse{\boolean{showcommentsMA}}{
  \newcommand*{\VN}[2][]{\MyCommentMA[#1]{blue}{VN}{#2}}
}{%
  \newcommand*{\VN}[2][]{}
}

\begin{document}

\newacronym{cnn}{CNN}{Convolutional Neural Network}
\newacronym{hr}{HR}{high-resolution}
\newacronym{lp}{LP}{License Plate}
\newacronym{lpr}{LPR}{License Plate Recognition}
\newacronym{lr}{LR}{low-resolution}
\newacronym{misr}{MISR}{Multi-Image Super-Resolution}
\newacronym{mse}{MSE}{Mean Squared Error}
\newacronym{ocr}{OCR}{Optical Character Recognition}
\newacronym{psnr}{PSNR}{Peak Signal-to-Noise Ratio}
\newacronym{sisr}{SISR}{Single-Image Super-Resolution}
\newacronym{sr}{SR}{super-resolution}
\newacronym{srcnn}{SRCNN}{Super-Resolution Convolutional Neural Network}
\newacronym{ssim}{SSIM}{Structural Similarity Index Measure}
\newacronym{vsr}{VSR}{Video Super-Resolution}
\newacronym{rcb}{RCB}{Residual Concatenation Block}
\newacronym{fm}{FM}{Feature Module}
\newacronym{sfe}{SFE}{Shallow Feature Extractor}
\newacronym{rm}{RM}{Reconstruction Module}
\newacronym{ps}{PS}{\textit{PixelShuffle}}
\newacronym{pu}{PU}{\textit{PixelUnshuffle}}
\newacronym{nn}{NN}{Neural Network}
\newacronym{psfe}{PSFE}{Pre-shallow Feature Extractor}
\newacronym{alpr}{ALPR}{Automatic License Plate Recognition}
\newacronym{ca}{CA}{Channel Unit}
\newacronym{pos}{POS}{Positional Unit}
\newacronym{tfam}{TFAM}{Two-fold Attention Module}
\newacronym{map}{MAP}{Maximum a Posteriori}
\newacronym{gan}{GAN}{Generative Adversarial Network}
\newacronym{ccpd}{CCPD}{Chinese City Parking Dataset}
\newacronym{mprnet}{MPRNet}{Multi-Path Residual Network}
\newacronym{cbam}{CBAM}{Convolution Block Attention Module}
\newacronym{se}{SE}{Squeeze-and-excitation}
\newacronym{esa}{ESA}{Enhanced Spatial Attention}
\newacronym{csrgan}{CSRGAN}{Character-Based Super-Resolution Generative Adversarial Networks}
\newacronym{dconv}{DConv}{depthwise-separable convolutional layer}
\newacronym{gp}{GP}{Geometrical Perception Unit}
\newacronym{srgan}{SRGAN}{Super-Resolution Generative Adversarial Networks}
\newacronym{dpca}{DPCA}{Dual-Coordinate Direction Perception Attention}
\newacronym{dganesr}{D\textunderscore GAN\textunderscore ESR}{Double Generative Adversarial Networks for Image Enhancement and Super Resolution}
\newacronym{lcofl}{LCOFL}{Layout and Character Oriented Focal Loss}
\newacronym{gplpr}{GP\_LPR}{Global Perception License Plate Recognition}

\newacronym{rdb}{RDB}{Residual Dense Block}
\newacronym{pltfam}{PLTFAM}{Pixel Level Three-Fold Attention Module}
\newacronym{lpd}{LPD}{License Plate Detection}

\newacronym{csbc}{CSBC}{Brazilian Computer Society Congress}
\newacronym{gt}{GT}{Ground Truth}
\newacronym{gpm}{GPM}{Global Perception Module}
\newacronym{dsam}{DSAM}{Deformable Spatial Attention Module}
\newacronym{hc}{HC}{Highest Confidence}
\newacronym{mv}{MV}{Majority Vote}
\newacronym{mvcp}{MVCP}{Majority Vote by Character Position}

\newacronym{plnet}{PLNET}{Pixel-Level Network}
\newacronym{edsr}{EDSR}{Enhanced Deep Super-Resolution Network}
\newacronym{ecbsr}{ECBSR}{Edge-oriented Convolution Block Super-Resolution}
\newacronym{lcdnet}{LCDNet}{Layout-Aware and Character-Driven Network}
\newacronym{ecb}{ECB}{Edge-oriented Convolution Block}
\newacronym{realesrgan}{Real-ESRGAN}{Real Enhanced Super-Resolution Generative Adversarial Networks}
\newacronym{lpsrgan}{LPSRGAN}{License Plate Super-Resolution Generative Adversarial Networks}
\newacronym{sr3}{SR3}{Super-Resolution via Iterative Refinement}
\newacronym{fps}{FPS}{frames per second}

\newcommand{\rodosolalpr}{RodoSol-ALPR\xspace}
\newcommand{\ufpralpr}{UFPR-ALPR\xspace}
\newcommand{\srplates}{UFPR-SR-Plates\xspace} %
\newcommand{\dataset}{\srplates}

\newcommand{\iwpod}{IWPOD-NET\xspace}

\newcommand{\ocrchina}{GP\_LPR\xspace}

\begin{frontmatter}
\maketitle

\begin{mail}
Department of Informatics, Federal University of Paraná, R. Evaristo F. Ferreira da Costa 391, Jardim das Américas, Curitiba, PR, 81530-090, Brazil. 
\end{mail}

\begin{dates}

\ifarxiv
\else
\small{\textbf{Received:} DD Month YYYY~~~$\bullet$~~~\textbf{Accepted:} DD Month YYYY~~~$\bullet$~~~\textbf{Published:} DD Month YYYY}
\fi

\end{dates}

\begin{abstract} 
\textbf{Abstract.} 
\noindent Recent advancements in super-resolution for \gls*{lpr} have sought to address challenges posed by \gls*{lr} and degraded images in surveillance, traffic monitoring, and forensic applications.
However, existing studies have relied on private datasets and simplistic degradation models.
To address this gap, we introduce \dataset, a novel dataset containing 10,000 tracks with 100,000 paired low and high-resolution license plate images captured under real-world conditions.
We establish a benchmark using multiple sequential \gls*{lr} and \gls*{hr} images per vehicle --~five of each~-- and two state-of-the-art models for super-resolution of license plates.
We also investigate three fusion strategies to evaluate how combining predictions from a leading \gls*{ocr} model for multiple super-resolved license plates enhances overall performance.
Our findings demonstrate that super-resolution significantly boosts LPR performance, with further improvements observed when applying majority vote-based fusion techniques.
Specifically, the \gls*{lcdnet} model combined with the \gls*{mvcp} strategy led to the highest recognition rates, increasing from $1.7$\% with low-resolution images to $31.1$\% with super-resolution, and up to $44.7$\% when combining \gls*{ocr} outputs from five super-resolved images.
These findings underscore the critical role of super-resolution and temporal information in enhancing \gls*{lpr} accuracy under real-world, adverse conditions.
The proposed dataset is publicly available to support further research and can be accessed at: \textbf{\url{https://valfride.github.io/nascimento2024toward/}}.

\end{abstract}


\end{frontmatter}

\glsresetall
\newcommand{\cmmnt}[1]{}
\section{Introduction}
\label{sec:intro}

\gls*{lpr} systems have become increasingly popular across various practical applications, such as traffic monitoring and toll collection~\citep{laroca2021efficient, ke2023ultra, liu2024irregular}.
These systems are designed to accurately recognize characters on a \gls*{lp} after it has been detected within an image.

While recent studies on \gls*{lpr} have reported high recognition rates, the results are primarily based on experiments using \gls*{hr} images, where the \gls*{lp} characters are clearly defined and free from significant noise~\citep{silva2022flexible,laroca2023leveraging,rao2024license}.
However, accurately recognizing characters in \gls*{lr} or degraded images remains a significant~challenge.

In surveillance scenarios, images are often captured at low resolutions or are heavily compressed due to constraints in storage and bandwidth.
Hence, \gls*{lp} characters may become distorted, blend into the background, or overlap with neighboring characters, making recognition challenging.
This underscores the need for robust methods that can effectively handle these types of degradation.

Taking this into account, various image enhancement techniques, including \gls*{sr}, have been proposed to improve image quality~\citep{moussa2022forensic,nascimento2023super,nascimento2024superCTD,pan2024lpsrgan}.
Although these techniques aim to improve \gls*{lpr}, their performance is frequently assessed using metrics such as \gls*{ssim} and \gls*{psnr}, which are known not to correlate well with human assessment of visual quality or recognition accuracy~\citep{johnson2016perceptual,zhang2018unreasonable,mehri2021mprnet,liu2023blind}.
Furthermore, most studies relied solely on private datasets \citep{hamdi2021new, maier2022reliability, luo2024real}, hindering fair comparisons.
Many also adopted simplistic degradation models, where \gls*{lr} images were created by simply downsampling the original \gls*{hr}~images~\citep{nascimento2022combining, kim2024afanet, pan2024lpsrgan}.

In response to these limitations, we introduce a new publicly available dataset, \srplates\footnote{The \srplates dataset is publicly available at \url{https://valfride.github.io/nascimento2024toward/}.}.
It comprises $100{,}000$ \gls*{lp} images captured by a rolling shutter camera installed on a Brazilian road. 
\dataset includes $10{,}000$ \gls*{lp} tracks, each consisting of ten consecutive images --~five \gls*{lr} images captured when the vehicle was farthest from the camera, and five \gls*{hr} images captured at its closest point.
These tracks, recorded under varying environmental and lighting conditions, feature two distinct \gls*{lp} layouts: Brazilian and Mercosur.
Unlike synthetically generated datasets, \dataset offers a more accurate representation of surveillance scenarios and makes it a valuable resource for advancing \gls*{lp} super-resolution research.

In summary, the main contributions of this work are:

\begin{itemize}
    \item \major{A publicly available dataset containing $100{,}000$ images, divided into $10{,}000$ tracks, with each track containing five \gls*{lr} images and five \gls*{hr} images of the same \gls*{lp}.
    To enhance variability, $5{,}000$ tracks were collected at a resolution of $1280\times960$ pixels, while the remaining $5{,}000$ tracks were captured at $1920\times1080$ pixels.
    The dataset is evenly distributed between Mercosur and Brazilian \glspl*{lp}, making it the largest dataset in terms of the number of \glspl*{lp} for both layouts;}
    \item \major{We conducted benchmark experiments on the proposed dataset using five state-of-the-art super-resolution models: (i) general-purpose approaches (\acrshort*{sr3}~\citep{saharia2023image}, \acrshort*{realesrgan}~\citep{wang2021real}), and (ii) \gls*{lp}-specialized networks (\acrshort*{lpsrgan}~\citep{pan2024lpsrgan}, \acrshort*{plnet}~\citep{nascimento2024superCTD}, and \acrshort*{lcdnet}~\citep{nascimento2024enhancing}).
    For each track, we generated five super-resolved images from the \gls*{lr} images and compared the recognition results obtained by the leading \gls*{ocr} model, \acrshort*{gplpr}~\citep{liu2024irregular}.
    The super-resolution process significantly boosted recognition accuracy, increasing from $2.2$\% to $29.9$\% for a single super-resolved image.
    To further enhance \gls*{lpr} performance, we explored three fusion strategies for combining the outputs from the \gls*{ocr} model based on multiple super-resolved images.
    Notably, applying the \gls*{mvcp} strategy with five super-resolved images improved the recognition rate from $29.9\%$ to $42.3\%$. The proposed dataset enables the exploration of temporal relationships among low-resolution \glspl*{lp}, as it includes multiple sequential \gls*{lr} images for each~\gls*{lp}.}

\end{itemize}

The remainder of this paper is structured as follows.
\cref{sec:RelatedWork} provides a brief review of related works.
In \cref{sec:ProposedApproach}, we introduce the \dataset dataset.
The experiments are detailed in \cref{sec:Experiments}. 
Finally, \cref{sec:Conclusions} concludes the paper by summarizing our findings and their significance.

\section{Related Work}
\label{sec:RelatedWork}

\glsreset{gplpr}
\glsreset{plnet}
\glsreset{lcdnet}

This section provides an overview of relevant works in \gls*{lpr} and \gls*{lp} super-resolution.
More specifically, \cref{rw:lpr} covers recent advancements and techniques in \gls*{lpr}, highlighting key innovations and their impact on recognition accuracy.
\cref{rw:srlpr} discusses the integration of super-resolution methods with \gls*{lpr}, focusing on their role in improving the recognition of low-quality or degraded \gls*{lp} images.

\subsection{
License Plate Recognition (LPR)
}
\label{rw:lpr}

The primary goal of \gls*{lpr} is to accurately identify characters from a given \gls*{lp} image.
To tackle this challenge, \citet{silva2020realtime} proposed treating the \gls*{lpr} stage as an object detection task, where each character class is identified as a distinct object.
They introduced CR-NET, a model based on YOLO~\citep{redmon2016yolo}, which has shown significant effectiveness for \gls*{lpr} in subsequent studies \citep{laroca2021efficient,oliveira2021vehicle,silva2022flexible}.

Recently, advancements in the field have moved toward holistic treatment of the entire \gls*{lp} image for text recognition, departing from traditional segmentation methods that isolate individual characters.
This shift has improved recognition accuracy while also enhancing computational efficiency. For instance, \cite{ke2023ultra} introduced a lightweight, multi-scale \gls*{lpr} network that integrates global channel attention layers to effectively fuse low- and high-level features.

To address real-world challenges, such as substantially tilted \glspl*{lp} caused by suboptimal camera positioning, recent research has focused on incorporating attention mechanisms into deep learning models.
\cite{rao2024license} integrated attention mechanisms into a CRNN model, while \cite{liu2024improving} introduced deformable spatial attention modules to enhance feature extraction and capture the \gls*{lp}'s global layout.
Building on this, \cite{liu2024irregular} presented a robust \gls*{ocr} model called \gls*{gplpr} for recognizing irregular LPs through the use of deformable spatial attention and global perception modules (this model is further detailed in \cref{xp:models}).
These advancements collectively address challenges such as attention deviation and character misidentification, leading to state-of-the-art performance on popular datasets such as CCPD~\citep{xu2018towards} and \rodosolalpr~\citep{laroca2022cross}.

Although these models have reported impressive accuracy and inference speed, most evaluations were conducted on datasets where all \gls*{lp} characters are clearly legible, even on challenging scenarios involving tilted \glspl*{lp}.
This setup, however, does not accurately reflect real-world surveillance environments, where cost-effective cameras are typically used and bandwidth limitations often degrade image quality.
Consequently, \gls*{lr} images with blurry characters that blend into adjacent ones and the \gls*{lp} background are prevalent, posing a substantial challenge for robust \gls*{lpr}~\citep{moussa2022forensic, ke2023ultra,schirrmacher2023benchmarking}.

\subsection{Super-Resolution for LPR}
\label{rw:srlpr}

The quality of an image is affected by various factors, including lighting, weather, camera distance, motion blur, and storage techniques, each introducing unique noise patterns.
These factors, combined with the structural variability in low-resolution \glspl*{lp}, make it challenging for \gls*{lpr} systems to accurately identify characters in such images.
Although recent advancements in \gls*{sr} methods have shown potential in improving character visibility in low-quality images~\citep{liu2023blind},
the specific challenges related to \gls*{lpr} under degraded conditions remain largely unaddressed~\citep{maier2022reliability, hijji2023intelligent, angelika2024yolov8}. 

To address these challenges, \cite{lin2021license} proposed an ESRGAN-based~\citep{wang2019esrgan} approach for \gls*{lp} image enhancement called PatchGAN.
This method leverages a residual dense network with progressive upsampling to preserve high-frequency details effectively.
While PatchGAN achieved impressive \gls*{psnr} and \gls*{ssim} values, its performance gains over other \gls*{gan}-based methods (e.g., SRGAN~\citep{ledig2017photo}) were relatively modest, and the model's complexity may limit its applicability in real-time scenarios. 

Expanding on this, \cite{hamdi2021new} proposed an approach called Double Generative Adversarial Networks for Image Enhancement and Super-Resolution, which employs two sequential networks: the first to deblur the image and the second to apply super-resolution, yielding the final output.
While their approach demonstrated effectiveness, it was only tested on synthetically generated low-resolution images created from high-resolution~\glspl*{lp}.

Recognizing the gap in integrating character recognition into the \gls*{sr} process, \cite{lee2020super} introduced a perceptual loss based on features extracted from scene text recognition models. 
Specifically, they leveraged intermediate representations from ASTER \citep{shi2019aster} to train a model based on \glspl*{gan}.
Their experiments showed that adding this perceptual loss improved results compared to models trained without it.
However, a lack of detailed information on datasets and degradation methods hinders the reproducibility and generalization of their approach.

Building on these developments, \cite{pan2023super} introduced a complete pipeline for the \gls*{sr} of \glspl*{lp} followed by recognition, utilizing ESRGAN~\citep{wang2019esrgan} for single-character enhancement.
Their method demonstrated effectiveness with moderately low-quality \glspl*{lp}, nevertheless, it struggled with severely degraded images, particularly when character boundaries were unclear.
To overcome these challenges, \cite{pan2024lpsrgan} developed \acrshort*{lpsrgan}, which processes the entire \gls*{lp} image and incorporates a degradation model that generates more realistic low-resolution \glspl*{lp}.
Despite these improvements, \acrshort*{lpsrgan} still struggled to reconstruct characters under severely degraded~conditions. 

\cite{kim2024afanet} introduced AFA-Net, an architecture that integrates deblurring sub-networks at both the pixel and feature levels.
Their method was evaluated on a dataset containing low-resolution and blurred \gls*{lp} images captured from unconstrained dash cams.
While the results were promising, the \gls*{lr} images were artificially generated through simple interpolation, and the testing protocol focused solely on super-resolving digits, excluding letter recognition.
This limits the generalizability of their approach to real-world \gls*{lp}~images.

\cite{luo2024real} designed a domain-specific degradation model to simulate real-world \gls*{lp} degradations, incorporating common factors such as motion blur, lighting issues, and noise.
By retraining ESRGAN on a dataset of high-resolution \glspl*{lp} with these simulated degradations, they achieved robust recognition performance.
This success highlights the model's effectiveness in controlled settings where degradation types are aligned with the training data.
However, when confronted with unconstrained real-world scenarios, with severe occlusion, extreme lighting variations, or unforeseen blur, the method struggled to preserve accurate character structure, indicating that further refinement is necessary for application in complex, real-world~settings.

\cite{alhalawani2024diffplate} recently introduced DiffPlate, a diffusion model for \gls*{lp} super-resolution that outperformed ESRGAN and SwinIR~\citep{liang2021swinir} in terms of \gls*{psnr} and \gls*{ssim}. 
However, its high computational cost restricts its real-time applicability in surveillance systems.
Furthermore, the model was trained and tested using synthetically generated images, where high-resolution \glspl*{lp} were downsampled by a factor of four to produce low-resolution~counterparts.

\cite{nascimento2023super,nascimento2024superCTD} introduced the \gls*{plnet}, a super-resolution model that employs specialized attention modules to enhance the quality of \gls*{lp} images.
While \gls*{plnet} showed potential in improving character clarity in \gls*{lr} scenarios, the experiments were limited to synthetically generated \gls*{lp} images.
In subsequent research, the same authors~\citep{nascimento2024enhancing} developed the \gls*{lcdnet}, which further improved character structure and positioning in \gls*{lp} layouts through deformable convolutions, shared-weight attention modules, and a GAN-based approach with an \gls*{ocr} discriminator and a layout-aware perceptual loss.
Although this latter study included preliminary experiments with real-world images, the dataset was not made publicly available, which limits reproducibility.
Both PLNET and LCDNet are described in more detail in \cref{xp:models}, as they are utilized in our~experiments.

\major{\cite{sendjasni2024embedding} proposed RDASRNet, a framework designed for extreme license plate \gls*{sr} (with a \(\times16\) scaling factor).
The architecture integrates a hierarchical channel attention mechanism that iteratively refines features extracted from \gls*{lr} inputs.
To enhance training, a dual-loss strategy was employed --~combining mean squared error with a contrastive loss guided by a Siamese network.
This approach enforces perceptual and structural consistency between the super-resolved outputs and \gls*{hr} ground-truth images within a latent space.
While RDASRNet achieves state-of-the-art performance on synthetic benchmarks, its high computational cost poses challenges for real-time deployment in surveillance systems. 
Furthermore, its reliance on synthetic training data --~where \gls{lr} images are generated via idealized downsampling~-- raises concerns about its generalizability to real-world degradations, a limitation shared with other studies~\citep{pan2023super, alhalawani2024diffplate}.}

In summary, while substantial progress has been made in super-resolution techniques for \gls*{lpr}, a major barrier to further advancement is the limited availability of public datasets containing paired low- and high-resolution \glspl*{lp}.
Most existing research relies on either proprietary datasets \citep{lee2020super, hamdi2021new, maier2022reliability, pan2024lpsrgan} or synthetically generated \gls*{lr} images \citep{pan2023super, alhalawani2024diffplate, kim2024afanet, luo2024real, sendjasni2024embedding}.

\begin{figure*}[!htb]

    \centering

    \resizebox{0.995\linewidth}{!}{
    \includegraphics[width=0.15\linewidth, height=0.096\linewidth]{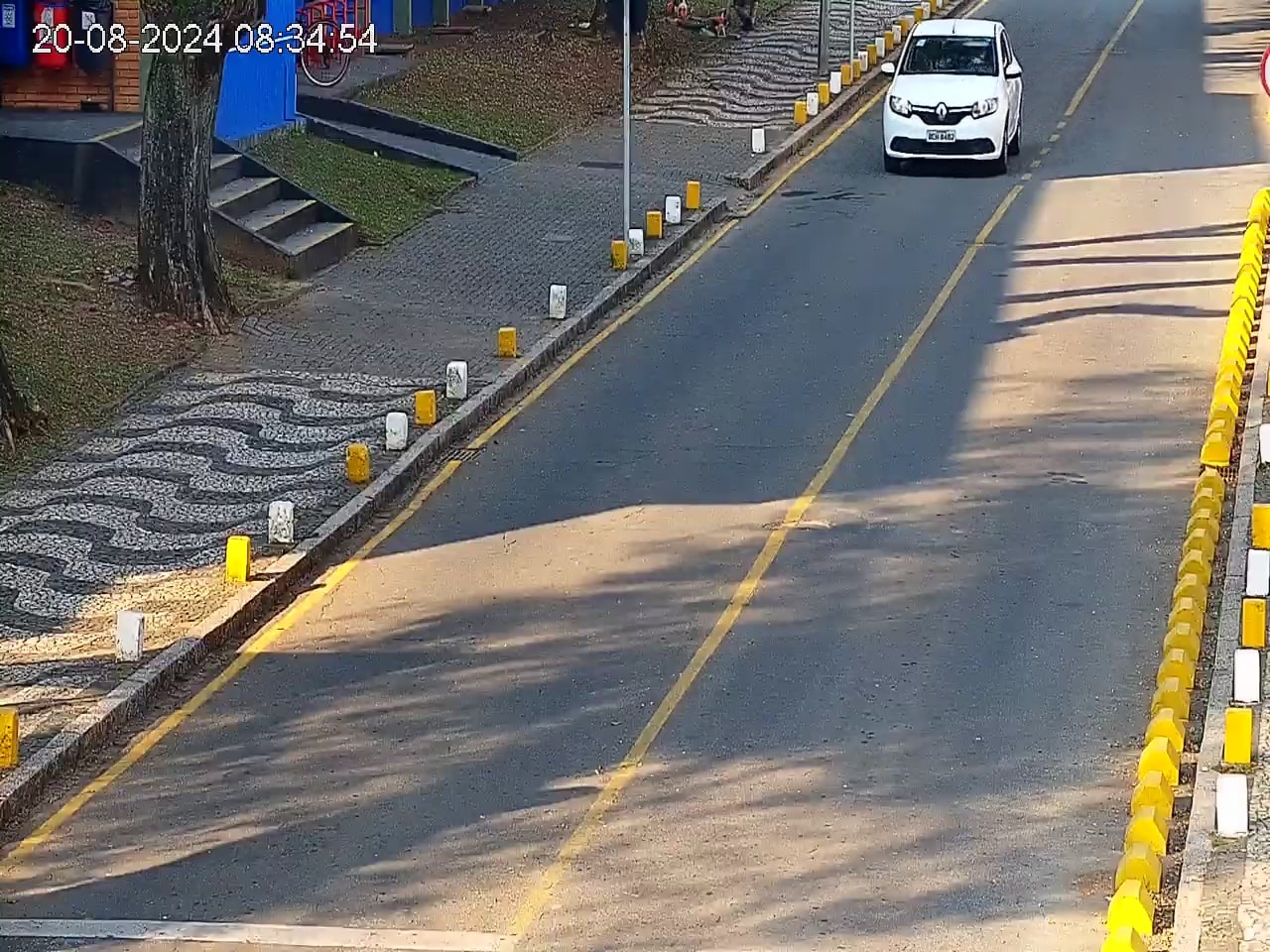}\hspace{-0.5mm} 
    \includegraphics[width=0.15\textwidth, height=0.096\textwidth]{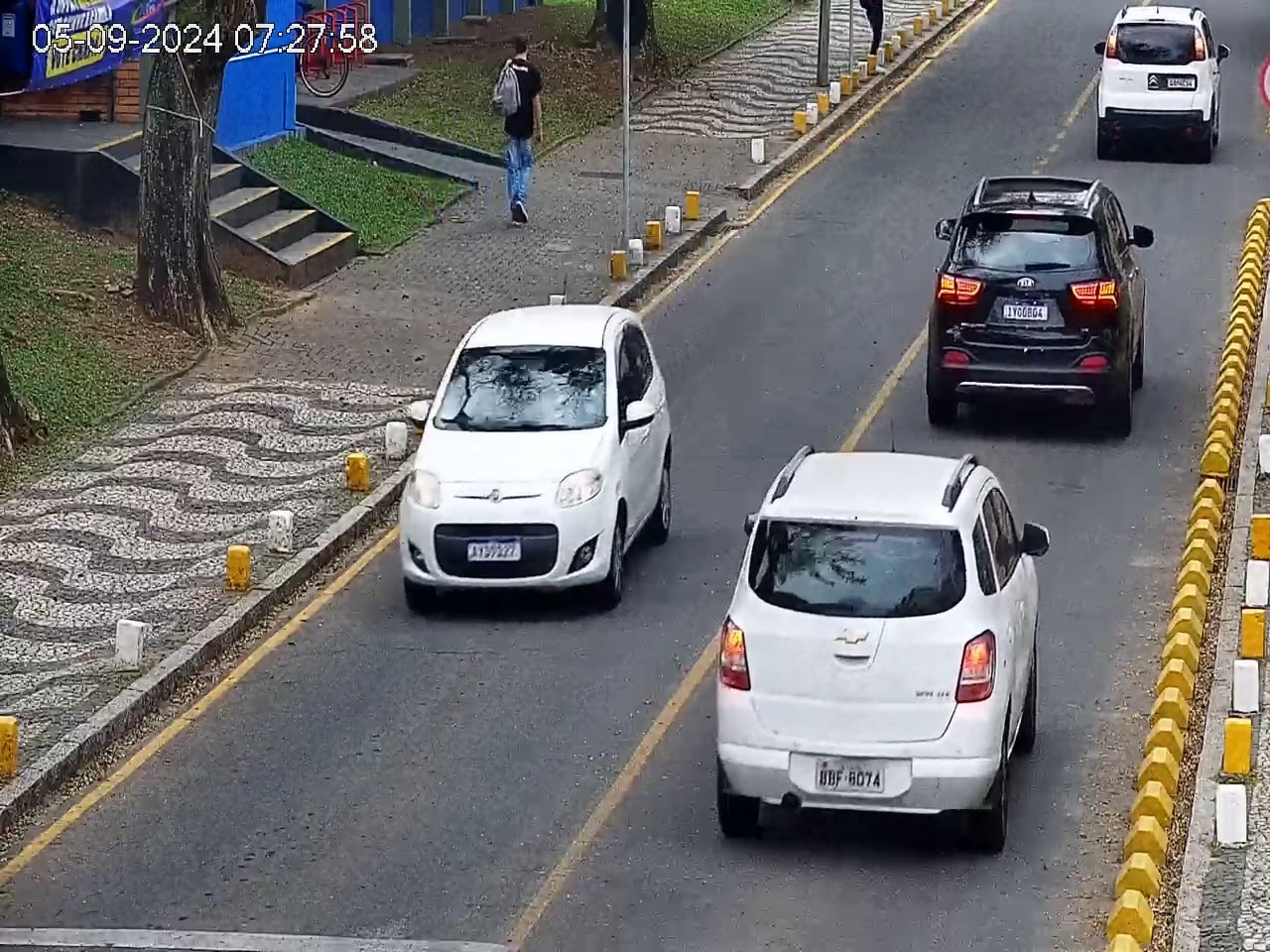}\hspace{-0.5mm}
    \includegraphics[width=0.15\textwidth, height=0.096\textwidth]{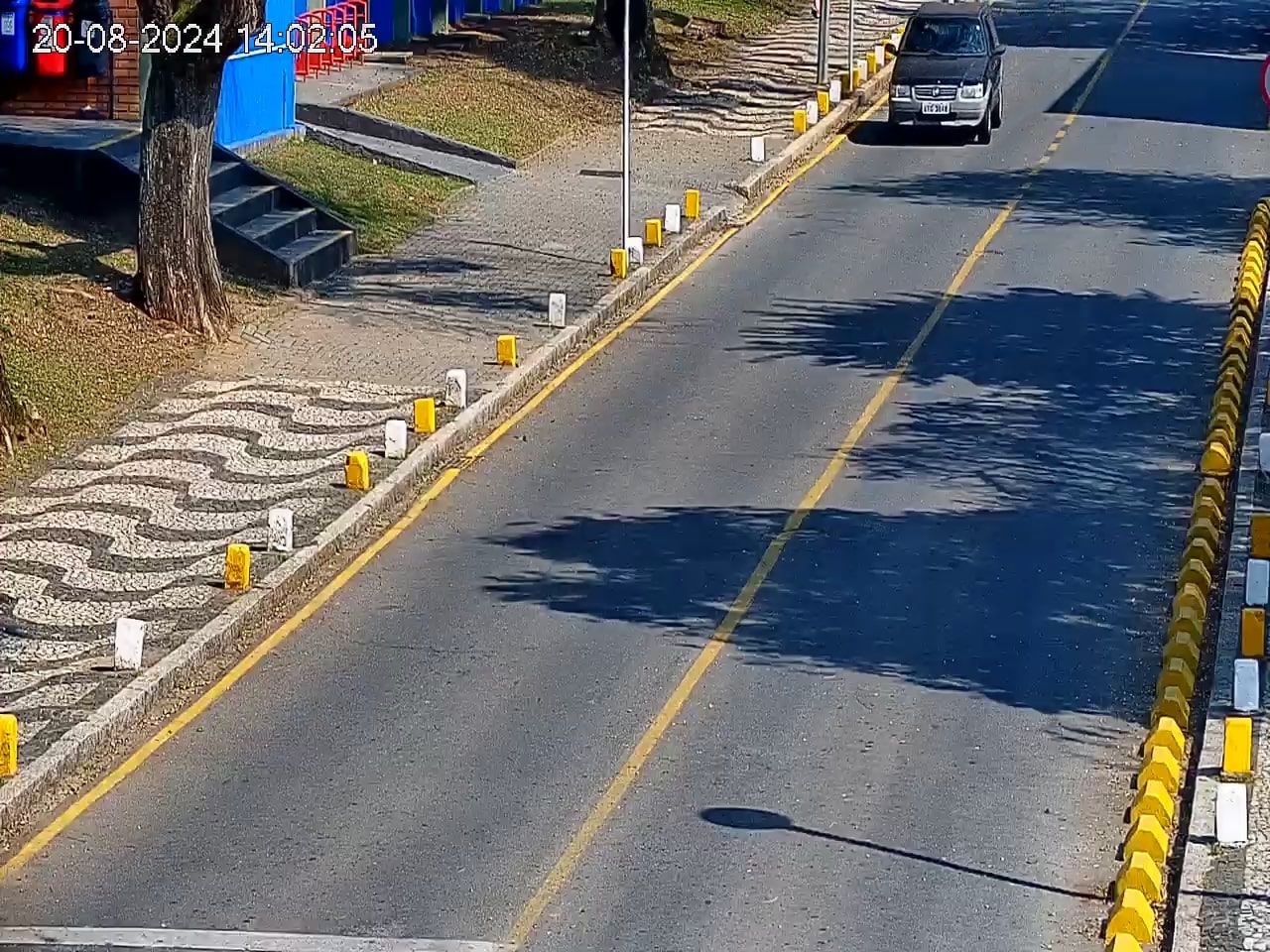}\hspace{-0.5mm}
    \includegraphics[width=0.15\textwidth, height=0.096\textwidth]{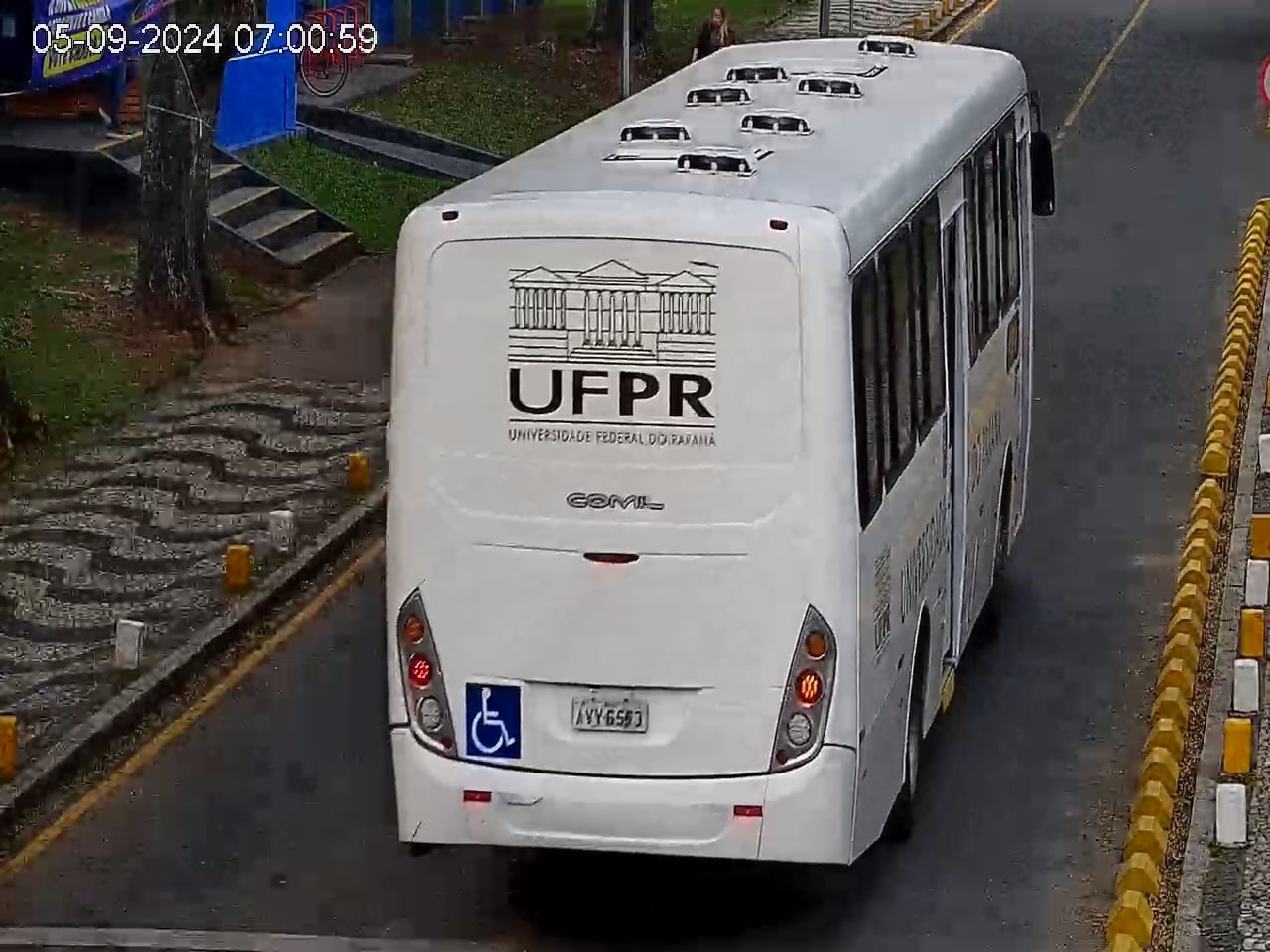}
    }
    
    \vspace{0.15mm}

    \resizebox{0.995\linewidth}{!}{
    \includegraphics[width=0.15\textwidth, height=0.096\textwidth]{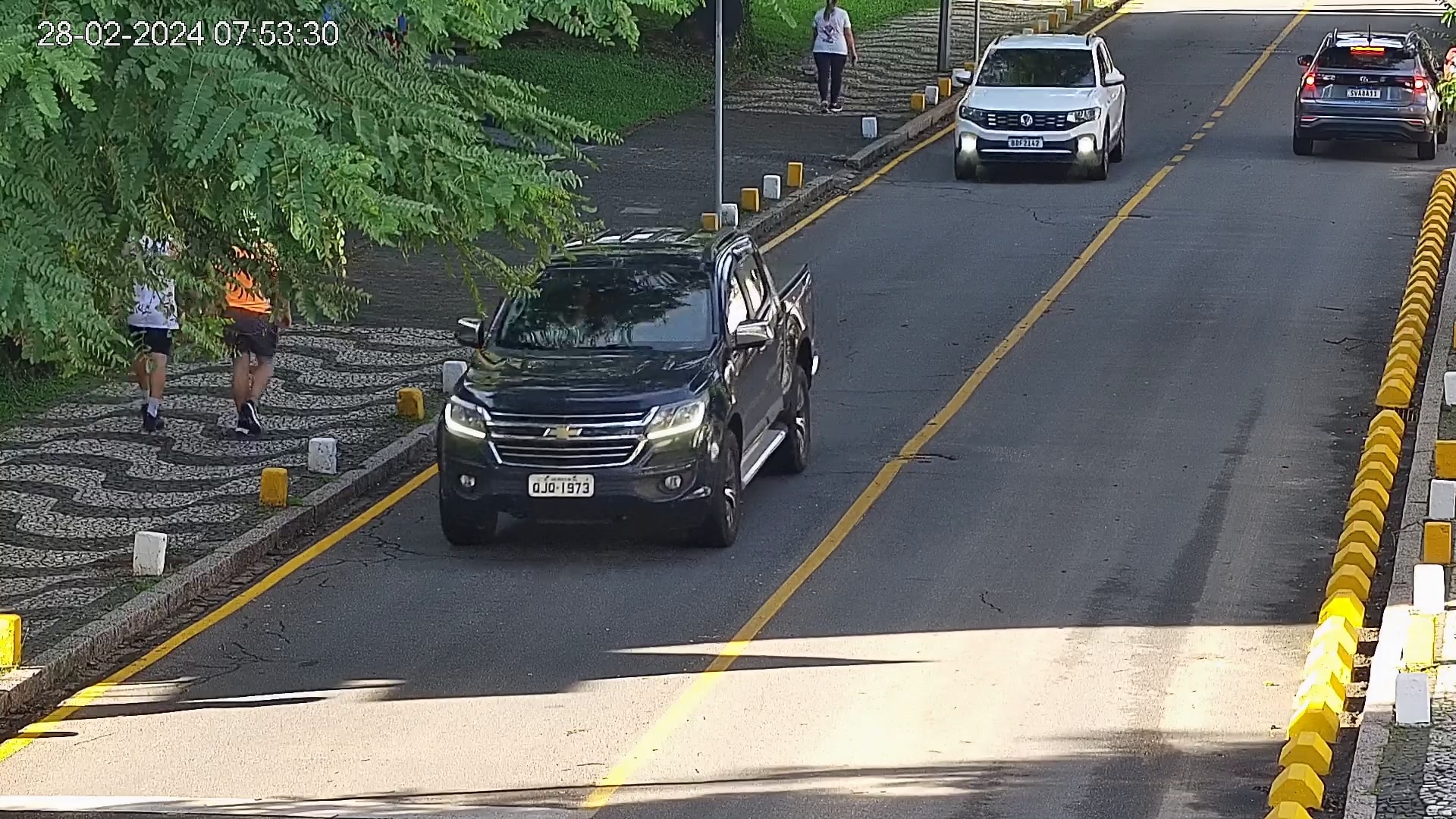}\hspace{-0.5mm} 
    \includegraphics[width=0.15\textwidth, height=0.096\textwidth]{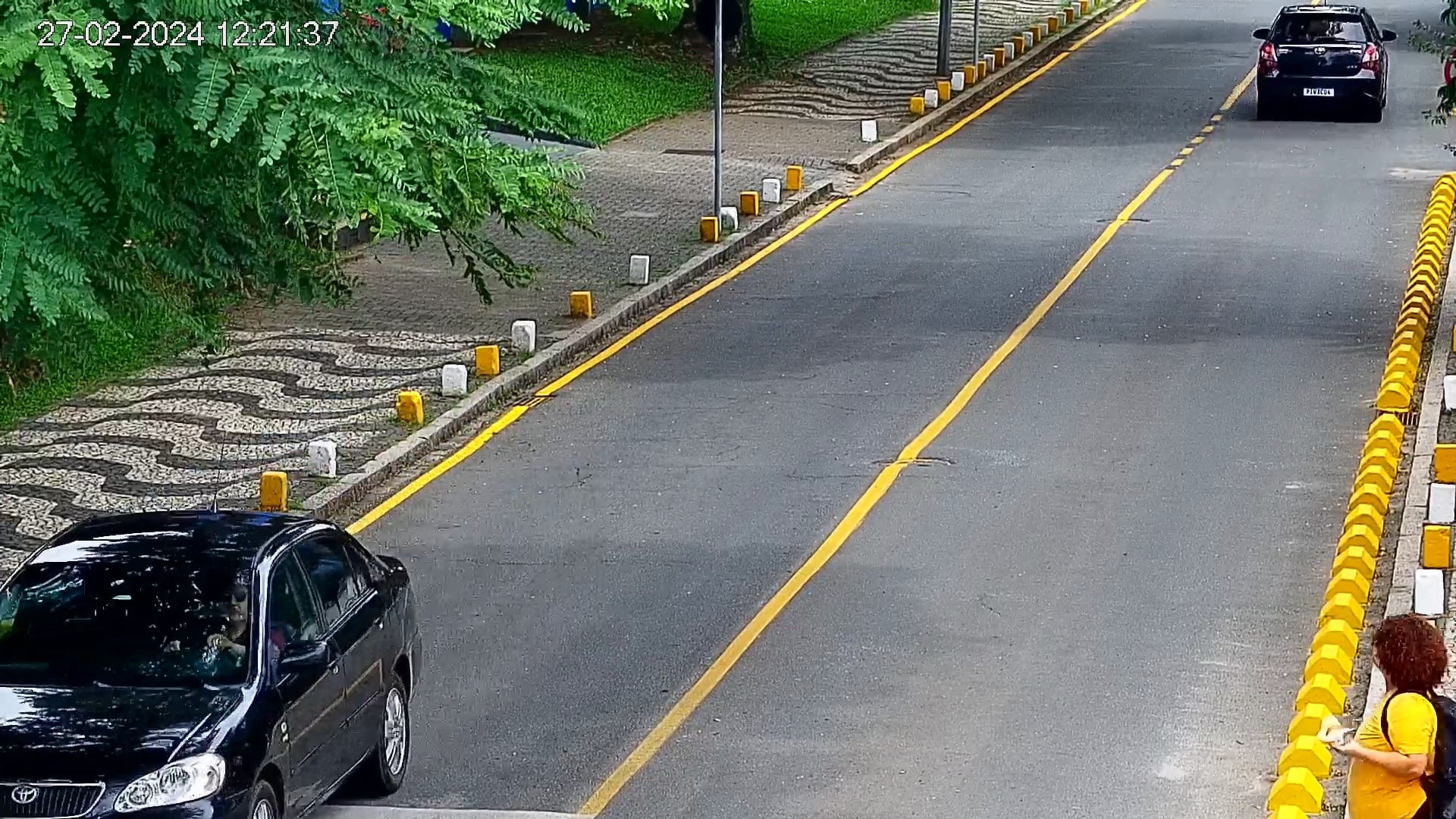}\hspace{-0.5mm}
    \includegraphics[width=0.15\textwidth, height=0.096\textwidth]{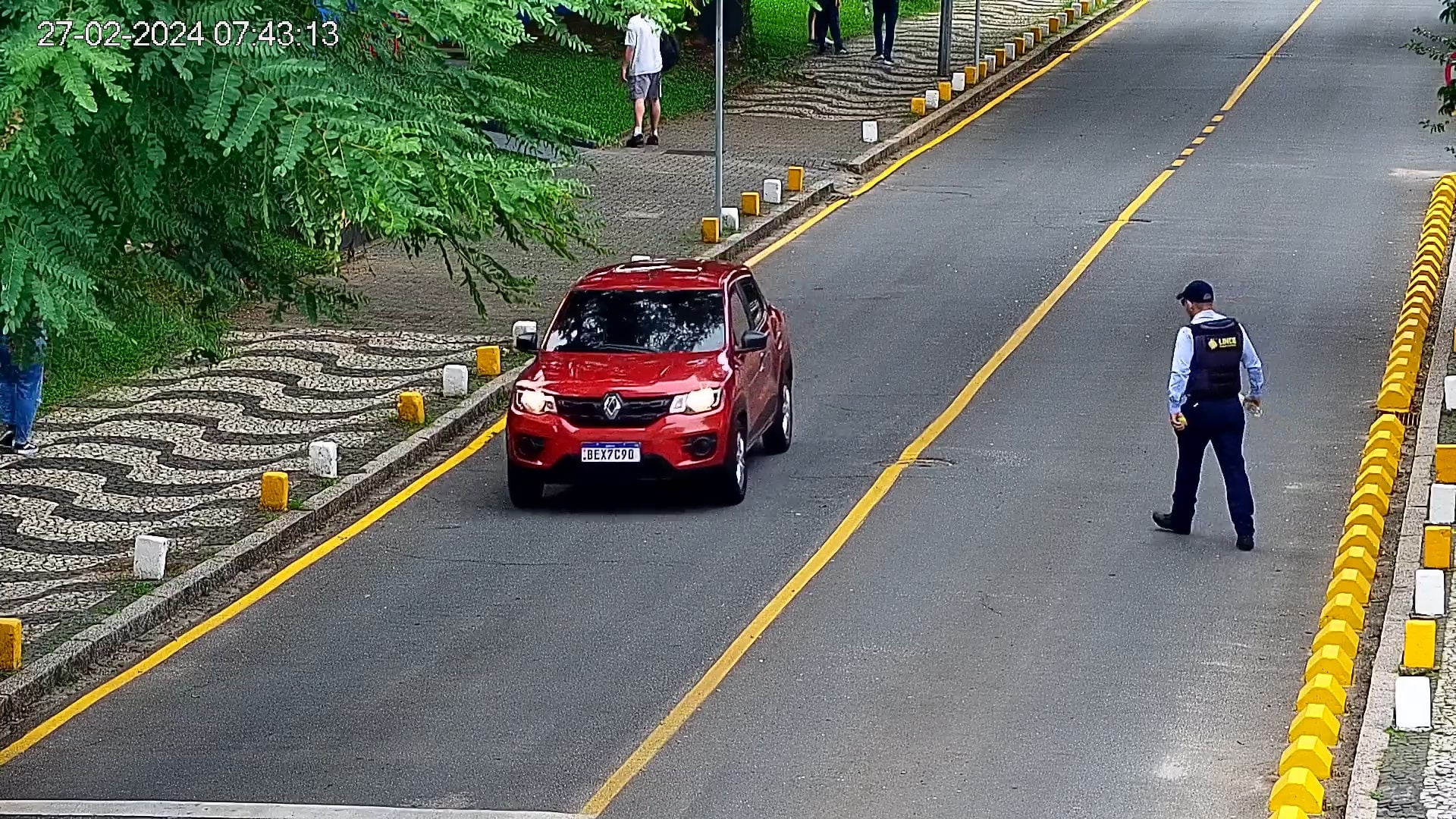}\hspace{-0.5mm}
    \includegraphics[width=0.15\textwidth, height=0.096\textwidth]{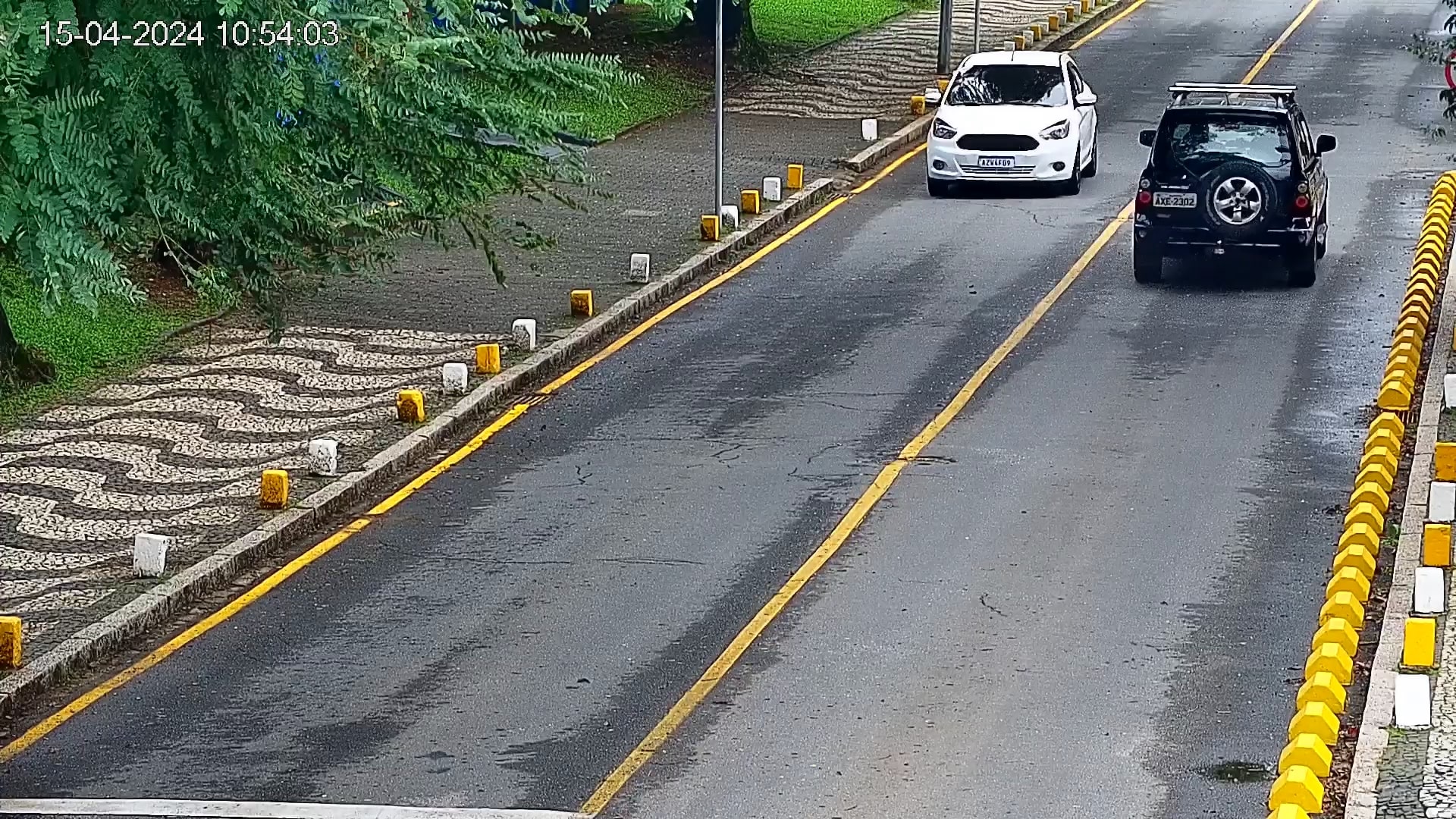}
    }

    \vspace{0.6mm}
    
    \caption{
    Examples of scenarios from which the \gls*{lp} images in the \dataset dataset were extracted. These images showcase a variety of vehicle types and their corresponding \glspl*{lp}, captured under different environmental conditions. The first row shows images taken with a resolution of $1280\times960$ pixels, while the second row displays images captured at $1920\times1080$ pixels. For better visualization, all images in this figure were slightly~resized.}
    
    \label{fig:original-samples}
\end{figure*}

\section{The \dataset Dataset}
\label{sec:ProposedApproach}

The \dataset dataset comprises $10{,}000$ tracks, each with five consecutive \gls*{lr} images and five consecutive \gls*{hr} images of the same \gls*{lp}.
With a total of $100{,}000$ images, this dataset is well-suited for \gls*{sr} and \gls*{lpr} tasks.
It offers real-world images captured under diverse noise and degradation conditions, while enabling direct \gls*{lr}-\gls*{hr} comparison and analysis of temporal variations within frame~sequences.

\newcommand{\vs}{\vspace*{-2.1mm}}
\newcommand{\hs}{\hspace*{-1.35mm}}
\begin{figure*}[!hbt]
    \captionsetup[subfigure]{labelformat=empty,position=top,captionskip=0.75pt,justification=centering} 
    
    \vspace{0.7mm}
    
    \centering

    \resizebox{0.997\linewidth}{!}{
    \subfloat[]{
    \includegraphics[width=0.15\textwidth, height=0.06\textwidth]{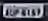}
    }\hs
    \subfloat[]{
    \includegraphics[width=0.15\textwidth, height=0.06\textwidth]{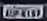}
    }\hs
    \subfloat[\large LR Images]{
    \includegraphics[width=0.15\textwidth, height=0.06\textwidth]{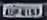}
    }\hs
    \subfloat[]{
    \includegraphics[width=0.15\textwidth, height=0.06\textwidth]{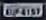}
    }\hs
    \subfloat[]{
    \includegraphics[width=0.15\textwidth, height=0.06\textwidth]{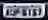}
    }\hspace{1mm}
    \subfloat[]{
    \includegraphics[width=0.15\textwidth, height=0.06\textwidth]{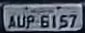}
    }\hs
    \subfloat[]{
    \includegraphics[width=0.15\textwidth, height=0.06\textwidth]{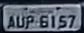}
    }\hs
    \subfloat[\large HR Images]{
    \includegraphics[width=0.15\textwidth, height=0.06\textwidth]{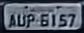}
    }\hs
    \subfloat[]{
    \includegraphics[width=0.15\textwidth, height=0.06\textwidth]{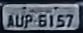}
    }\hs
    \subfloat[]{
    \includegraphics[width=0.15\textwidth, height=0.06\textwidth]{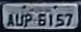}
    }\hs
    }

    \vs

    \resizebox{0.997\linewidth}{!}{
    \subfloat[]{
    \includegraphics[width=0.15\textwidth, height=0.06\textwidth]{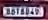}
    }\hs
    \subfloat[]{
    \includegraphics[width=0.15\textwidth, height=0.06\textwidth]{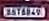}
    }\hs
    \subfloat[]{
    \includegraphics[width=0.15\textwidth, height=0.06\textwidth]{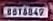}
    }\hs
    \subfloat[]{
    \includegraphics[width=0.15\textwidth, height=0.06\textwidth]{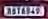}
    }\hs
    \subfloat[]{
    \includegraphics[width=0.15\textwidth, height=0.06\textwidth]{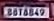}
    }\hspace{1mm}
    \subfloat[]{
    \includegraphics[width=0.15\textwidth, height=0.06\textwidth]{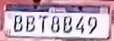}
    }\hs
    \subfloat[]{
    \includegraphics[width=0.15\textwidth, height=0.06\textwidth]{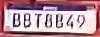}
    }\hs
    \subfloat[]{
    \includegraphics[width=0.15\textwidth, height=0.06\textwidth]{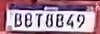}
    }\hs
    \subfloat[]{
    \includegraphics[width=0.15\textwidth, height=0.06\textwidth]{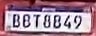}
    }\hs
    \subfloat[]{
    \includegraphics[width=0.15\textwidth, height=0.06\textwidth]{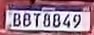}
    }\hs
    }

    \vs

    \resizebox{0.997\linewidth}{!}{
    \subfloat[]{
    \includegraphics[width=0.15\textwidth, height=0.06\textwidth]{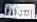}
    }\par\hs
    \subfloat[]{
    \includegraphics[width=0.15\textwidth, height=0.06\textwidth]{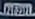}
    }\hs
    \subfloat[]{
    \includegraphics[width=0.15\textwidth, height=0.06\textwidth]{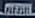}
    }\hs
    \subfloat[]{
    \includegraphics[width=0.15\textwidth, height=0.06\textwidth]{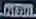}
    }\hs
    \subfloat[]{
    \includegraphics[width=0.15\textwidth, height=0.06\textwidth]{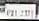}
    }\hspace{1mm}
    \subfloat[]{
    \includegraphics[width=0.15\textwidth, height=0.06\textwidth]{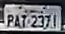}
    }\hs
    \subfloat[]{
    \includegraphics[width=0.15\textwidth, height=0.06\textwidth]{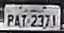}
    }\hs
    \subfloat[]{
    \includegraphics[width=0.15\textwidth, height=0.06\textwidth]{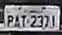}
    }\hs
    \subfloat[]{
    \includegraphics[width=0.15\textwidth, height=0.06\textwidth]{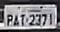}
    }\hs
    \subfloat[]{
    \includegraphics[width=0.15\textwidth, height=0.06\textwidth]{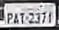}
    }\hs
    }

    \vs

    \resizebox{0.997\linewidth}{!}{
    \subfloat[]{
    \includegraphics[width=0.15\textwidth, height=0.06\textwidth]{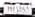}
    }\hs
    \subfloat[]{
    \includegraphics[width=0.15\textwidth, height=0.06\textwidth]{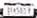}
    }\hs
    \subfloat[]{
    \includegraphics[width=0.15\textwidth, height=0.06\textwidth]{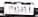}
    }\hs
    \subfloat[]{
    \includegraphics[width=0.15\textwidth, height=0.06\textwidth]{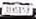}
    }\hs
    \subfloat[]{
    \includegraphics[width=0.15\textwidth, height=0.06\textwidth]{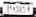}
    }\hspace{1mm}
    \subfloat[]{
    \includegraphics[width=0.15\textwidth, height=0.06\textwidth]{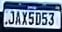}
    }\hs
    \subfloat[]{
    \includegraphics[width=0.15\textwidth, height=0.06\textwidth]{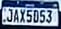}
    }\hs
    \subfloat[]{
    \includegraphics[width=0.15\textwidth, height=0.06\textwidth]{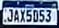}
    }\hs
    \subfloat[]{
    \includegraphics[width=0.15\textwidth, height=0.06\textwidth]{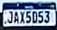}
    }\hs
    \subfloat[]{
    \includegraphics[width=0.15\textwidth, height=0.06\textwidth]{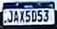}
    }\hs
    }
    
    \vspace{0.6mm}

    \caption{Examples of tracks from the \dataset dataset. Each track comprises five consecutive \gls*{lr} images and five consecutive \gls*{hr} images of the same \gls*{lp}, captured under varying conditions.
    Each row shows a single track, with the \gls*{lr} images displayed on the left and the corresponding \gls*{hr} images on the right. We remark that even what we consider `\gls*{hr}' in the context of this work is of lower quality than the datasets typically used in \gls*{lpr} research.}
    \label{fig:tracks-extracted}
\end{figure*}

The images were taken with a rolling shutter camera near the Department of Informatics at the Federal University of Paraná in Curitiba, Brazil, simulating real-world surveillance conditions.
\cref{fig:original-samples} presents sample images collected to build the \dataset dataset (i.e., prior to \gls*{lp} detection and extraction), highlighting the diversity of vehicle types, including cars, buses, and trucks.

Data collection occurred over six months, with images captured daily during a fixed 10-hour interval, from 7~a.m. to 5~p.m.
Nighttime images were excluded due to infrared interference, which often caused overexposure or underexposure, making \gls*{lp} recognition challenging even in \gls*{hr}~images.

The original images were evenly distributed between two video resolutions: $1280 \times 960$ pixels and $1920 \times 1080$ pixels. 
This variation in resolution enables the extraction of \glspl*{lp} with different pixel densities, allowing for a broader evaluation of \gls*{sr} methods across varying image quality.
Consequently, \dataset becomes more versatile for tasks that require adaptation to different levels of detail.

The proposed dataset is also balanced between two \gls*{lp} layouts: Brazilian and Mercosur.
Brazilian \glspl*{lp} follows a format of three letters followed by four digits, while Mercosur \glspl*{lp} features three letters, one digit, one letter, and two digits.
Although both types of \glspl*{lp} are similar in size and shape, they differ significantly in color schemes and character~fonts.

\begin{figure*}[!htb]
    \centering
    \resizebox{0.85\linewidth}{!}{
    \includegraphics[width=\linewidth]{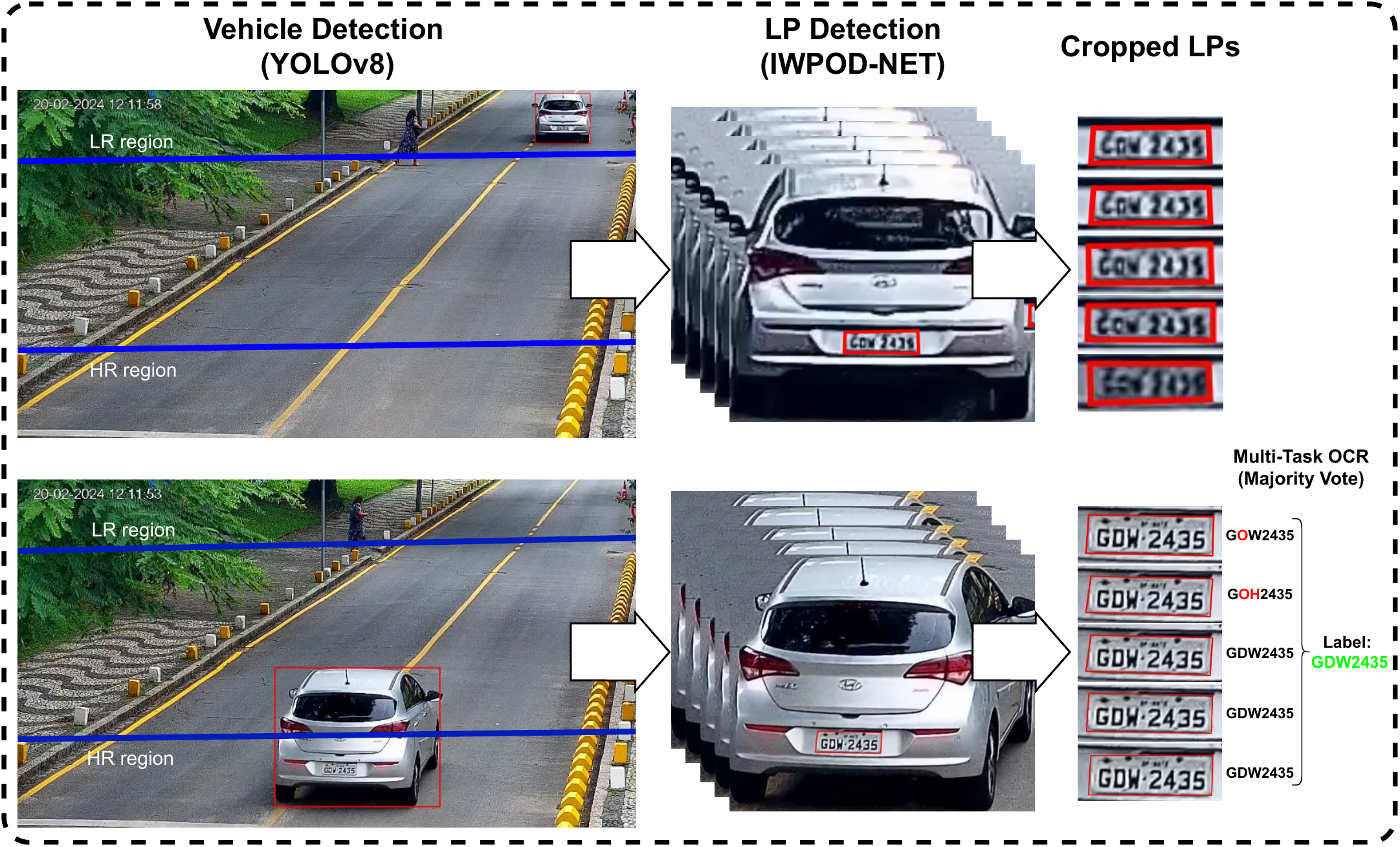}
    }

    \vspace{0.6mm}
    
    \caption{Illustration of the process of vehicle identification and tracking using \textit{YOLOv8}~\citep{yolov8}, followed by \gls*{lp} corner detection with \iwpod~\citep{silva2022flexible}.
    LP images were extracted from the original frames based on the detected corners.
    A multi-task \gls*{ocr} model, proposed by \cite{goncalves2018realtime}, was applied to recognize the text on the extracted LP images. Majority voting was applied to determine the final label.
    The blue lines in the original images demarcate the regions for detecting low-resolution \glspl*{lp} (above the higher line) and high-resolution \glspl*{lp} (below the lower line).
    }
    \label{fig:pipeline}
\end{figure*}

As shown in \cref{fig:pipeline}, vehicle tracks were derived from video footage capturing vehicles entering and exiting the road from opposite sides. 
To detect and track vehicles, we employed the YOLOv8 model~\citep{yolov8}, a choice motivated by the proven effectiveness of the YOLO family in object detection in unconstrained scenarios~\citep{lima2024toward,laroca2021efficient,laroca2025improving}.
We fine-tuned the model to meet our specific needs, starting with a pre-trained version and collecting initial bounding box annotations for detected vehicles.
After each detection round, we carefully reviewed and manually corrected any annotation errors, incorporating these adjustments into the training set and retraining the model.
This iterative process progressively improved detection accuracy, culminating in a refined dataset of $2{,}954$ labeled images tailored to our~application.

We cropped each vehicle's region of interest using the annotations from the process described above.
Subsequently, we applied \iwpod~\citep{silva2022flexible} to locate the \gls*{lp} corners.
Although \iwpod is a well-regarded method for this task~\citep{jia2023efficient,wei2024efficient}, its original training on high-quality images limited its effectiveness in our dataset, particularly for vehicles at greater distances.
To overcome this limitation, we retrained \iwpod from scratch with optimized hyperparameters, significantly improving its robustness in detecting \glspl*{lp} from distant~vehicles.
More specifically, we started with an initial set of $300$ annotated \glspl*{lp} to train the \iwpod model.
Through an iterative process, we tested the model on new images, corrected any errors, and progressively expanded the training set with these refined annotations.
After several iterations, we conducted a final training phase with $839$ images, ensuring precise \gls*{lp} corner detection considering our scenario.
We then applied the model to extract \glspl*{lp} using a minimum bounding box method, adding 20\% padding to both vertical and horizontal dimensions to capture contextual~surroundings.

From the resulting sequences, we selected the five \gls*{lr} images that were farthest from the camera.
To annotate the \gls*{lp} characters, we employed the multi-task \gls*{ocr} developed by \cite{goncalves2018realtime} on each of the five \gls*{hr} samples in the track, utilizing a sequence-level majority vote strategy.
\cref{fig:tracks-extracted} shows all \gls*{lp} images from five tracks in the~dataset.

Each image within a track is also accompanied by a JSON file containing the coordinates $(x,~y)$ of its four corners, the layout of the \gls*{lp} (Brazilian or Mercosur), and its textual content (e.g., ABC-1234).
\cref{xp:statisticsLps} presents a summary of key statistical characteristics for each layout across both resolutions in the \dataset dataset, including the median, maximum, and minimum dimensions, as well as the total number of unique \glspl*{lp}. 
\label{xp:stats}

\begin{table}[!htb]
\centering
\caption{Summary of statistics (in pixels) for Brazilian and Mercosur \glspl*{lp} across resolutions in the \dataset dataset. 
}
\label{xp:statisticsLps}
\resizebox{\linewidth}{!}{
\begin{tabular}{lcclcclcclcc} 
\toprule
\multicolumn{1}{c}{\multirow{3}{*}{\dataset}} & \multicolumn{5}{c}{$1280\times960$} & & \multicolumn{5}{c}{$1920\times1080$} \\ 
\cmidrule{2-6} \cmidrule{8-12}
\multicolumn{1}{c}{} & \multicolumn{2}{c}{Brazilian} &  & \multicolumn{2}{c}{Mercosur} &  & \multicolumn{2}{c}{Brazilian} &  & \multicolumn{2}{c}{Mercosur} \\
\multicolumn{1}{c}{} & LR & HR &  & LR & HR &  & LR & HR &  & LR & HR \\ 
\midrule
Median Height & $\phantom{0}19$ & $\phantom{0}34$ &  & $\phantom{0}18$ & $\phantom{0}32$ &  & $21$ & $\phantom{0}50$ &  & $\phantom{0}21$ & $\phantom{0}38$ \\
Median Width  & $\phantom{0}35$ & $\phantom{0}69$ &  & $\phantom{0}35$ & $\phantom{0}67$ &  & $49$ & $100$           &  & $\phantom{0}49$ & $100$           \\
Max Height    & $\phantom{0}28$ & $\phantom{0}60$ &  & $\phantom{0}25$ & $\phantom{0}49$ &  & $28$ & $\phantom{0}52$ &  & $\phantom{0}26$ & $\phantom{0}52$ \\
Min Height    & $\phantom{0}15$ & $\phantom{0}22$ &  & $\phantom{0}13$ & $\phantom{0}23$ &  &$14$  & $\phantom{0}26$ &  & $\phantom{0}16$ & $\phantom{0}26$ \\
Max Width     & $\phantom{0}56$ & $103$           &  & $\phantom{0}59$ & $\phantom{0}88$ &  & $35$ & $122$           &  & $\phantom{0}61$ & $122$           \\
Min Width     & $\phantom{0}26$ & $\phantom{0}51$ &  & $\phantom{0}24$ & $\phantom{0}50$ &  &$34$ & $\phantom{0}71$ &  & $\phantom{0}34$ & $\phantom{0}71$  \\

\midrule
Unique \glspl*{lp}  & \multicolumn{2}{c}{$1{,}663$} &  & \multicolumn{2}{c}{$2{,}500$} &  & \multicolumn{2}{c}{$1{,}627$} &  & \multicolumn{2}{c}{$2{,}496$}\\
\bottomrule
\end{tabular}
}
\end{table}

Although the dataset acquisition process was automated, all annotations were manually reviewed to ensure the reliability of the \dataset dataset for research purposes.
Despite the popularity of the chosen \gls*{ocr} model in the literature~\citep{goncalves2019multitask,nascimento2024enhancing}, we found that it produced errors in approximately $5$\% of the tracks.
These errors were rectified during the aforementioned analysis~process.

For the experimental protocol, we divided each resolution set ($1920 \times 1080$ and $1280 \times 960$ pixels) into approximately $40$\%, $20$\%, and $40$\% for training, validation, and testing, respectively.
This resulted in $3{,}965$ tracks for training, $2{,}030$ for validation, and $4{,}005$ for testing. 
Note that the number of tracks in the training, validation, and test sets does not conform to the exact ratios of $4{,}000$/$2{,}000$/$4{,}000$ images, as we carefully avoided any overlap of \glspl*{lp} across the training, validation, and test sets, even when the same vehicle/\gls*{lp} was depicted in images of different resolutions. 
For instance, if the \gls*{lp} ``ABC-1234'' is included in the training set, it is strictly excluded from both the test and validation sets. 
As demonstrated by \cite{laroca2023do}, the presence of near-duplicate \gls*{lp} images across different subsets can artificially inflate model performance.
By preventing such overlaps, \dataset creates a fair and reliable resource for training, validating, and testing deep learning-based~models.

Regarding privacy concerns, \glspl*{lp} of vehicles registered in Brazil are not associated with the personal information of the vehicle owner, thus mitigating the risk of privacy breaches.
Each \gls*{lp} uniquely identifies the vehicle itself~\citep{brlaw,oliveira2021vehicle}.

\section{Experimental Results}
\label{sec:Experiments}

This section delves into the experimental details of this work.
\cref{xp:models} introduces the models employed, providing an overview of the framework and key hyperparameters.
\cref{xp:fusioning} outlines the fusion strategies implemented to combine the predictions generated by the \gls*{ocr} model for multiple super-resolved images, aiming to enhance \gls*{lpr} performance.
Finally, \cref{xp:results} presents and discusses the~results.

All experiments were conducted on a computer equipped with an AMD Ryzen $5950$X $3.4$ GHz CPU, $128$ GB of RAM, an SSD with read speeds of $535$ MB/s and write speeds of $445$ MB/s, and an NVIDIA Quadro RTX~$8000$ GPU~($48$~GB).

\subsection{Models}
\label{xp:models}

\major{We applied the \gls*{gplpr} model~\citep{liu2024irregular} to \gls*{lpr} on super-resolved images generated from the \gls*{lr} images in the \dataset dataset. 
The \gls*{sr} process was conducted using five state-of-the-art networks: 
(i)~two general-purpose models: \acrshort*{realesrgan}~\citep{wang2021real} and SR3~\citep{saharia2023image},
and (ii)~three \gls*{lp}-specialized models \acrshort*{lpsrgan}~\citep{pan2024lpsrgan}, \gls*{plnet}~\citep{nascimento2023super,nascimento2024superCTD}, and \gls*{lcdnet}~\citep{nascimento2024enhancing}.}

\major{The chosen models achieved state-of-the-art performance in both general \gls*{sr} tasks~\citep{saharia2023image, luo2024real} and \gls*{lp}-specific \gls*{sr} tasks~\citep{nascimento2023super, pan2024lpsrgan}.
Due to the absence of an official implementation, we reimplemented \acrshort*{lpsrgan} based on the methodology described by \cite{pan2024lpsrgan}.}
\major{The code for all models 
is publicly available\footnote{\gls*{gplpr}: \url{https://github.com/mmm2024/gp_lpr/}}$^,$\footnote{\gls*{plnet}: \url{https://github.com/valfride/lpr-rsr-ext/}}$^,$\footnote{\gls*{lcdnet}: \url{https://github.com/valfride/lpsr-lacd/}}$^,$\footnote{\acrshort*{realesrgan}: \url{https://github.com/xinntao/Real-ESRGAN/}}$^,$\footnote{\acrshort*{lpsrgan}: \url{https://github.com/valfride/lpsrgan/}}}.

The \gls*{gplpr} model focuses on \gls*{lpr} of irregular LPs, employing attention mechanisms to handle perspective distortion.
Central to its architecture is the global perception module, which enhances character feature completeness by incorporating global visual information. This facilitates global interaction among features, distinguishing characters with similar structures and minimizing misidentifications. Additionally, the model employs the Deformable Spatial Attention Module, featuring deformable convolution layers that adjust to variations in character positions and shapes, improving the network's ability to capture the overall layout of \glspl*{lp}.

\major{Real-ESRGAN~\citep{wang2021real} extends ESRGAN~\citep{wang2018esrgan} by introducing a high-order degradation model tailored for real-world scenarios. Unlike traditional approaches, it trains exclusively on synthetically degraded data generated through a rigorous pipeline that applies multiple degradation steps --~including blurring, noise injection, resizing, and compression. A key innovation lies in its artifact suppression mechanism, which employs filtering to mitigate distortions introduced during degradation simulation. 
The framework adopts a U-Net discriminator with spectral normalization to stabilize adversarial training. Training proceeds in two stages: first, a \gls*{psnr}-oriented optimization ensures structural fidelity, followed by a perceptual refinement stage to enhance visual quality. This methodology has demonstrated state-of-the-art performance in general image restoration, validating the effectiveness of synthetic degradation modeling for practical super-resolution tasks.} 

\major{\acrshort*{sr3}~\citep{saharia2023image} takes a distinct approach by framing super-resolution as a diffusion process. Unlike common GAN-based methods, \gls*{sr3} iteratively refines a noisy input image into a high-resolution output through a Markov chain. This stochastic refinement enables the model to explore multiple plausible reconstructions, particularly advantageous for recovering fine-grained details in severely degraded \gls*{lp} images. The iterative process is guided by a noise prediction network trained to reverse a predefined degradation schedule, making \gls*{sr3} robust to diverse noise types and compression artifacts. 
}

\major{LPSRGAN~\citep{pan2024lpsrgan} enhances \gls*{lpr} accuracy in unconstrained scenarios through a three-pronged approach. It introduces an n-stage random combination degradation (n-RCD) model to simulate real-world degradations like blur, noise, and compression via multi-stage randomized combinations, addressing limitations of simplistic degradation pipelines. The framework adopts a modified RRDBNet+ generator, building on Residual-in-Residual Dense Blocks (RRDB)~\citep{wang2019esrgan} with dropout layers to improve feature representation and generalization across \gls*{lp} layouts. To prioritize character clarity for recognition systems, \acrshort*{lpsrgan} employs a perceptual loss optimized for \gls*{lpr}, aligning super-resolved images with a Connectionist Temporal Classification loss to optimize character clarity by aligning super-resolved images with \gls*{ocr} output predictions. This integration of realistic degradation modeling, architectural enhancements, and task-specific optimization enables robust restoration of degraded \glspl*{lp}, particularly under severe real-world distortions.}

The \gls*{plnet} model builds upon the foundation laid by~\cite{mehri2021mprnet}, introducing refinements specifically tailored for \gls*{lp} super-resolution.
It incorporates a shallow feature extractor module, using an autoencoder equipped with \textit{PixelShuffle} and \textit{PixelUnshuffle} layers~\citep{shi2016real}
to efficiently extract shallow features while preserving essential information through skip connections.
\gls*{plnet} also integrates a mechanism to capture inter-channel and spatial relationships, thus improving the model's ability to rearrange and scale input data more effectively than conventional methods.

\gls*{lcdnet} employs deformable convolutions and shared-weight attention modules within a \gls*{gan} framework.
An \gls*{ocr} model acts as the discriminator, steering the super-resolution process toward prioritizing \gls*{lpr} accuracy.
During training, \gls*{lcdnet} optimizes a loss function designed to preserve character structure and the overall integrity of the \gls*{lp} layout (e.g., penalizing any confusion between letters and digits).
This ensures both visually clear and accurate \gls*{lp}~recognition.

\major{Here we detail the key hyperparameters used for training the models, all implemented using the PyTorch framework.
The hyperparameters were selected based on the prior works~\citep{saharia2023image, pan2024lpsrgan, wang2021real, nascimento2024superCTD,nascimento2024enhancing, liu2024irregular} and preliminary experiments conducted on the validation set of the proposed dataset.
For all models, we employed the Adam optimizer. 
In \acrshort*{lpsrgan}~\citep{pan2024lpsrgan}, we replaced the original HyperLPR3 OCR with \cite{goncalves2019multitask}'s multi-task model trained on \srplates (ensuring fair comparison as HyperLPR3 was trained on Chinese plates with unavailable source code), using generator/discriminator learning rates of $10^{-4}$/$10^{-5}$ respectively.  
\acrshort*{realesrgan}~\citep{wang2021real} followed its standard two-stage protocol: PSNR-oriented pretraining for $10^6$ iterations ($lr=2{\times}10^{-4}$) followed by GAN fine-tuning for $4{\times}10^5$ iterations ($lr=1{\times}10^{-4}$), with EMA stabilization and U-Net discriminator. \gls*{sr3}~\citep{saharia2023image} employed 100 denoising steps across $10^6$ training iterations ($lr=1{\times}10^{-4}$), incorporating Gaussian blur augmentation on low-resolution inputs.}
For both \gls*{lcdnet} and \gls*{plnet}, the initial learning rate was set to $10$\textsuperscript{-$4$}, decaying by a factor of $0.8$ when no improvement in the loss function was observed.
For the \gls*{gplpr} model, the maximum decoding length was set to $K=7$ to match the $7$-character format of the \glspl*{lp} in the \srplates dataset.
To mitigate training oscillations in \gls*{gplpr}, following~\cite{liu2024irregular}, we applied a step learning rate scheduler starting with an initial learning rate of $10$\textsuperscript{-$3$} and decaying by a factor of $0.8$ every $50$~epochs. 

Additionally, we added padding with gray pixels to both the \gls*{lr} and \gls*{hr} images to preserve their aspect ratio before resizing them to dimensions of $16 \times 48$ and $32 \times 96$ pixels, respectively, corresponding to an upscale factor of~$2$.

\subsection{Fusion Methods}
\label{xp:fusioning}

\glsreset{hc}
\glsreset{mv}
\glsreset{mvcp}

\begin{figure*}[!htb]
    \centering
    \resizebox{0.65\linewidth}{!}{
    \includegraphics[width=\linewidth]{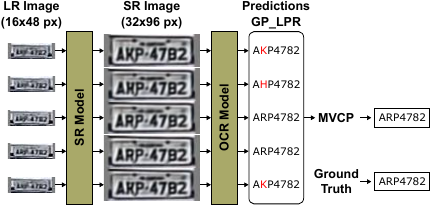}
    }

    \vspace{0.7mm}
    
    \caption{\major{Illustration of the fusion process to enhance \gls*{lpr} performance by combining multiple super-resolved \gls*{lp} images. 
    Sequential \gls*{lr} images (original size \(16\times48\) pixels) from each track are independently upsampled to \(32\times96\) pixels via a single-image \gls*{sr} model, then processed by the \gls*{gplpr} model~\citep{liu2024irregular}. 
    The \gls*{ocr} outputs are aggregated using three fusion strategies: \acrfull*{hc}, \acrfull*{mv}, and \acrfull*{mvcp}, leveraging temporal consistency across frames to resolve structural ambiguities in character reconstruction (e.g., distinguishing~``R'', ``K'' and~``H''). 
    This approach is particularly valuable for forensic and surveillance scenarios that demand high~reliability.}
    }
    \label{fig:experimentalSetup}
\end{figure*}

Inspired by the work of~\cite{laroca2023leveraging}, we examine three fusion methods to combine predictions from multiple super-resolved images generated by the \gls*{sr} networks. 
For each track in the test set, we process \(N\) consecutive \gls*{lr} images of the same \gls*{lp}, captured at varying distances from the camera. Each \gls*{lr} image is independently super-resolved and fed to the \gls*{ocr} model.
The predictions are then fused using one of the following strategies. In our experiments, \(N \in \{1, 3, 5\}\), with images selected sequentially from nearest to farthest. \major{This process is illustrated in \cref{fig:experimentalSetup}}.

\major{The first fusion method, \gls*{hc}, straightforwardly selects the single prediction with the highest associated confidence score as the final output:}
\[
      \hat{y} = y_{k},\text{ where } k = \arg\max_{i \in \{1,\ldots,N\}} P_i, \quad 
\]
\major{where \(P_i\) is the confidence score for the \(i\)-th \gls*{ocr} prediction~\(y_i\).
This strategy aligns with classical confidence-based fusion rules (e.g.,
 ~\citep{kittler1998combining}
), which prioritize predictions with the highest certainty.}

\major{The second method, \gls*{mv}, is an ensemble learning technique
~\citep{zhou2025ensemble}
that selects the most frequent prediction among all~\(N\)~outputs:}

\[
     \hat{y} = \text{mode}\left(\{y_1, y_2, \ldots, y_N\}\right), 
\]
for \(\text{mode}\) defined as:
\[
    \text{mode}(S) = \arg\max_{x \in S} \text{count}_{S}(x),
\]
where \(\text{count}_{S}(x)\) is the number of times \(x\) appears in the multiset \(S\).
\major{\gls*{mv} has been widely adopted in applications such as handwriting recognition
~\citep{zhao2020multiple}.}

\major{The third method, \gls*{mvcp}, aggregates predictions per character position:}
\[
    \hat{y}_j = \text{mode}\left(\{y_{(1,j)}, y_{(2,j)}, \ldots, y_{(N,j)}\}\right),
\]
\major{where \(y_{(i,j)}\) denotes the \(j\)-th character (for \(j \in [1, 7]\), \(j \in \mathbb{N}\)) of the \(i\)-th OCR prediction. The final output \(\hat{y}\) is formed by concatenating all \(\hat{y}_j\). 
This hierarchical approach draws inspiration from structured prediction frameworks~\citep{ghamrawi2005collective}, resolving ambiguities at each character~position.}

\major{A key challenge in majority-vote strategies is resolving ties between competing predictions. To address this, our approach prioritizes the prediction with the highest average character-level confidence score within tied groups.
Consider five \gls*{ocr} predictions for a Brazilian \gls*{lp} with associated confidence scores: two instances of ``ABC-1234'' ($0.95$ and $0.91$), two instances of ``ABD-1234'' ($0.89$ and $0.88$), and ``HBG-1284'' ($0.71$). Here, the \gls*{mv} method resolves the tie between the top two candidates (``ABC-1234'' and ``ABD-1234'') by comparing their confidence scores --~$0.93$ for ``ABC-1234'' and $0.88$ for ``ABD-1234''~-- selecting the former. For character-level ambiguities (e.g., a conflict between two ``C''s and two ``D''s in the third position), the \gls*{mvcp} strategy defaults to the character in the prediction with the highest confidence (in this case selecting ``C'' from ``ABC-1234'' with $0.95$ confidence).
This dual-layered approach ensures systematic tie-breaking while emphasizing statistically robust reconstructions through confidence metrics and character~prevalence.}

\major{These fusion strategies are adaptations of well-established methods in Pattern Recognition.
For instance, \gls*{mv}-based fusion has demonstrated effectiveness in enhancing \gls*{ocr} performance for degraded documents~\citep{reul2018improving}, while \gls*{hc} is a widely adopted approach in speech recognition systems~\citep{prabhavalkar2023end}.}

\subsection{Results}
\label{xp:results}

In this section, we present the results achieved by the \gls*{ocr} model in recognizing a single super-resolved \gls*{lp} image from each of the five \gls*{sr} baseline models. Following this, we analyze the employed fusion approaches and examine the impact of different resolutions on the results. Finally, we provide visual results for qualitative analysis.

In \gls*{lpr} research, model performance is traditionally evaluated using the recognition rate, defined as the ratio of correctly recognized \glspl*{lp} (where all characters are correctly classified) to the total number of \glspl*{lp} in the test set~\citep{wang2022rethinking,chen2023endtoend,wei2024efficient}.
In addition to this metric, we incorporate partial matches (cases where at least $5$ or $6$ characters are recognized correctly).
This approach is particularly valuable when not all \gls*{lp} characters can be accurately reconstructed or recognized, as it helps to narrow the search~space.

\major{\Cref{xp:quantitativeR1MJ} summarizes the performance of the \gls*{gplpr} model~\citep{liu2024irregular} on both low- and high-resolution \gls*{lp} images, alongside images super-resolved by general-purpose methods \acrshort*{sr3}~\citep{saharia2023image} and \acrshort*{realesrgan}~\citep{wang2021real}, as well as \gls*{lp}-specialized networks \acrshort*{lpsrgan}~\citep{pan2024lpsrgan}, \gls*{plnet}~\citep{nascimento2023super,nascimento2024superCTD}, and \gls*{lcdnet}~\citep{nascimento2024enhancing}.}
In \gls*{lr} samples, most characters are barely distinguishable, resulting in recognition rates as low as~$2.2$\%.

\begin{table}[!htb]
\centering
\caption{\major{Recognition rates obtained by \gls*{gplpr} on different test images. The first LR and HR images from each track were used for the experiments on the original images. Super-resolved images were generated from the first LR image of each~track.}}
\label{xp:quantitativeR1MJ}
\resizebox{0.85\linewidth}{!}{
\begin{tabular}{lccc} 
\toprule
 \multirow{2}{*}{Test Images}  & \multicolumn{3}{c}{\# Correct Characters}   \\ 
\cmidrule{2-4}
 \multicolumn{1}{c}{} & All    & $\ge6$ & $\ge5$                   \\ 
\midrule
 HR      &  $85.2\%$ & $98.5\%$  & $99.8\%$  \\
 LR (no SR)     &  $\phantom{0}2.2\%$ &  $\phantom{0}8.2\%$  & $20.1\%$   \\ 
\midrule
    LR + SR (SR3)    &  $18.4\% $   & $45.8\%$       & $68.3\%$   \\
    LR + SR (LPSRGAN)   &  $19.6\%$    & $46.1\%$     & $67.4\%$  \\
    LR + SR (Real-ESRGAN)   &  $20.2\%$    & $49.8\%$     & $71.8\%$  \\
    LR + SR (\gls*{plnet})    &  $29.9\% $   & $57.8\%$       & $76.9\%$   \\
    LR + SR (\gls*{lcdnet})   &  $29.9\%$    & $59.2\%$     & $77.1\%$  \\
\bottomrule
\end{tabular}
}
\end{table}

\begin{table*}[!htb]
\centering
\caption{
\major{Comparison of recognition rates achieved by combining the outputs of the \ocrchina model for multiple super-resolved images generated by \gls*{plnet} and \gls*{lcdnet}. As outlined in \cref{xp:fusioning}, three fusion strategies were evaluated: \acrfull*{hc}, \acrfull*{mv}, and \acrfull*{mvcp}.}}
\label{xp:quantitativeR35MJ}
\resizebox{0.975\linewidth}{!}{
\begin{tabular}{lcccccccccccc} 
\toprule
\multirow{2}{*}{\begin{tabular}[c]{@{}c@{}}Test Images\\(\textbf{Both Resolutions})\end{tabular}} & \multirow{2}{*}{\begin{tabular}[c]{@{}c@{}}\# Images \phantom{0} \\Majority Vote\end{tabular}} & \multicolumn{3}{c}{HC} &  & \multicolumn{3}{c}{MV} &  & \multicolumn{3}{c}{MVCP} \\ 
\cmidrule{3-5} \cmidrule{7-9} \cmidrule{11-13}
&  & All & $\geq 6$ & $\geq 5$ &  & All & $\geq 6$ & $\geq 5$ &  & All & $\geq 6$ & $\geq 5$ \\

\midrule 
\multicolumn{1}{l}{\multirow{2}{*}{\begin{tabular}[c]{@{}l@{}}SR3 (LR + SR) \end{tabular}}} & $3$ & \phantom{0}$19.9\%$ & \phantom{0}$49.5\%$ & \phantom{0}$70.8\%$ &  & \phantom{0}$21.4\%$ & \phantom{0}$50.9\%$ & \phantom{0}$71.2\%$ &  & \phantom{0}$24.1\%$ & \phantom{0}$56.3\%$ & \phantom{0}$76.4\%$ \\ 
\multicolumn{1}{l}{} & $5$ & \phantom{0}$20.2\%$ & \phantom{0}$50.6\%$ & \phantom{0}$71.9\%$ &  & \phantom{0}$25.1\%$ & \phantom{0}$55.1\%$ & \phantom{0}$73.8\%$ &  & \phantom{0}$28.3\%$ & \phantom{0}$61.8\%$ & \phantom{0}$81.5\%$ \\ 
\midrule
\multicolumn{1}{l}{\multirow{2}{*}{\begin{tabular}[c]{@{}l@{}}LPSRGAN (LR + SR) \end{tabular}}} & $3$ & \phantom{0}$24.0\%$ & \phantom{0}$51.5\%$ & \phantom{0}$70.7\%$ &  & \phantom{0}$24.5\%$ & \phantom{0}$51.7\%$ & \phantom{0}$70.9\%$ &  & \phantom{0}$25.3\%$ & \phantom{0}$53.5\%$ & \phantom{0}$72.9\%$ \\ 
\multicolumn{1}{l}{} & $5$ & \phantom{0}$25.4\%$ & \phantom{0}$53.1\%$ & \phantom{0}$71.9\%$ &  & \phantom{0}$27.4\%$ & \phantom{0}$54.3\%$ & \phantom{0}$73.0\%$ &  & \phantom{0}$28.8\%$ & \phantom{0}$56.3\%$ & \phantom{0}$76.9\%$ \\ 
\midrule
\multicolumn{1}{l}{\multirow{2}{*}{\begin{tabular}[c]{@{}l@{}}Real-ESRGAN (LR + SR) \end{tabular}}} & $3$ & \phantom{0}$23.5\%$ & \phantom{0}$54.6\%$ & \phantom{0}$76.0\%$ &  & \phantom{0}$24.1\%$ & \phantom{0}$55.8\%$ & \phantom{0}$76.2\%$ &  & \phantom{0}$25.6\%$ & \phantom{0}$57.4\%$ & \phantom{0}$78.3\%$ \\ 
\multicolumn{1}{l}{} & $5$ & \phantom{0}$24.9\%$ & \phantom{0}$56.6\%$ & \phantom{0}$77.6\%$ &  & \phantom{0}$27.7\%$ & \phantom{0}$59.4\%$ & \phantom{0}$78.5\%$ &  & \phantom{0}$29.5\%$ & \phantom{0}$61.7\%$ & \phantom{0}$81.6\%$ \\ 
\midrule 
\multirow{2}{*}{\begin{tabular}[c]{@{}l@{}}\gls*{plnet} (LR + SR) \end{tabular}} & $3$ & \phantom{0}$34.5\%$ & \phantom{0}$63.8\%$ & \phantom{0}$80.7\%$ &  & \phantom{0}$36.1\%$ & \phantom{0}$64.4\%$ & \phantom{0}$80.9\%$ &  & \phantom{0}$36.6\%$ & \phantom{0}$66.1\%$ & \phantom{0}$82.0\%$ \\
& $5$ & \phantom{0}$35.9\%$ & \phantom{0}$65.3\%$ & \phantom{0}$82.9\%$&  & \phantom{0}$39.5\%$ & \phantom{0}$67.4\%$ & \phantom{0}$83.6\%$ &  & \phantom{0}$40.9\%$ & \phantom{0}$70.4\%$ & \phantom{0}$85.8\%$ \\ 
\midrule
\multicolumn{1}{l}{\multirow{2}{*}{\begin{tabular}[c]{@{}l@{}}\gls*{lcdnet} (LR + SR) \end{tabular}}} & $3$ & \phantom{0}$34.8\%$ & \phantom{0}$63.6\%$ & \phantom{0}$81.9\%$ &  & \phantom{0}$36.7\%$ & \phantom{0}$64.5\%$ & \phantom{0}$82.0\%$ &  & \phantom{0}$37.8\%$ & \phantom{0}$67.0\%$ & \phantom{0}$83.3\%$ \\
\multicolumn{1}{l}{} & $5$ & \phantom{0}$35.7\%$ & \phantom{0}$64.8\%$ & \phantom{0}$83.1\%$ &  & \phantom{0}$40.7\%$ & \phantom{0}$68.1\%$ & \phantom{0}$84.2\%$ &  & \phantom{0}$42.3\%$ & \phantom{0}$72.0\%$ & \phantom{0}$86.9\%$ \\ 

\bottomrule
\end{tabular}
}
\end{table*}

\begin{table*}[!htb]
\centering
\caption{\major{Comparison of recognition rates achieved by combining the outputs of the \ocrchina model for multiple super-resolved images generated by PLNET and LCDNet (considering only \glspl*{lp} extracted from images with $1280\times960$ pixels). As outlined in \cref{xp:fusioning}, three fusion strategies were evaluated: \acrfull*{hc}, \acrfull*{mv}, and \acrfull*{mvcp}.}}
\label{xp:quantitativeR35MJ1280x960}
\resizebox{0.975\linewidth}{!}{
\begin{tabular}{lcccccccccccc} 
\toprule
\multirow{2}{*}{\begin{tabular}[c]{@{}c@{}}Test Images\\($\textbf{1280} \times \textbf{960}$)\end{tabular}} & \multirow{2}{*}{\begin{tabular}[c]{@{}c@{}}\# Images \phantom{0}\\Majority Vote\end{tabular}} & \multicolumn{3}{c}{HC} &  & \multicolumn{3}{c}{MV} &  & \multicolumn{3}{c}{MVCP}  \\ 
\cmidrule{3-5} \cmidrule{7-9} \cmidrule{11-13}
&  & All & $\geq 6$ & $\geq 5$ &  & All & $\geq 6$ & $\geq 5$ &  & All & $\geq 6$ & $\geq 5$ \\ 
\midrule
\multirow{2}{*}{SR3 (LR + SR)} & $3$ & \phantom{00}$9.5\%$ & \phantom{0}$32.2\%$ & \phantom{0}$56.8\%$ &  & \phantom{00}$9.8\%$ & \phantom{0}$33.3\%$ & \phantom{0}$57.2\%$ &  & \phantom{0}$11.8\%$ & \phantom{0}$38.9\%$ & \phantom{0}$63.9\%$ \\ 
& $5$ & \phantom{00}$9.9\%$ & \phantom{0}$34.0\%$ & \phantom{0}$58.4\%$ &  & \phantom{0}$12.1\%$ & \phantom{0}$37.1\%$ & \phantom{0}$60.3\%$ &  & \phantom{0}$15.3\%$ & \phantom{0}$46.4\%$ & \phantom{0}$71.6\%$ \\ 
\midrule
\multicolumn{1}{l}{\multirow{2}{*}{\begin{tabular}[c]{@{}l@{}}LPSRGAN (LR + SR) \end{tabular}}} & $3$ & \phantom{00}$8.7\%$ & \phantom{0}$30.3\%$ & \phantom{0}$52.2\%$ &  & \phantom{00}$8.9\%$ & \phantom{0}$30.3\%$ & \phantom{0}$52.4\%$ &  & \phantom{00}$8.7\%$ & \phantom{0}$31.1\%$ & \phantom{0}$55.9\%$ \\ 
\multicolumn{1}{l}{} & $5$ & \phantom{00}$9.6\%$ & \phantom{0}$31.1\%$ & \phantom{0}$54.0\%$ &  & \phantom{0}$10.2\%$ & \phantom{0}$32.3\%$ & \phantom{0}$55.6\%$ &  & \phantom{0}$11.0\%$ & \phantom{0}$34.6\%$ & \phantom{0}$60.9\%$ \\ 
\midrule
\multirow{2}{*}{Real-ESRGAN (LR + SR)} & $3$ & \phantom{0}$12.3\%$ & \phantom{0}$37.8\%$ & \phantom{0}$63.7\%$ &  & \phantom{0}$12.6\%$ & \phantom{0}$39.0\%$ & \phantom{0}$64.0\%$ &  & \phantom{0}$13.7\%$ & \phantom{0}$39.8\%$ & \phantom{0}$65.8\%$ \\ 
\multicolumn{1}{l}{} & $5$& \phantom{0}$14.0\%$ & \phantom{0}$40.4\%$ & \phantom{0}$65.6\%$ &  & \phantom{0}$15.7\%$ & \phantom{0}$42.5\%$ & \phantom{0}$66.8\%$ &  & \phantom{0}$17.4\%$ & \phantom{0}$45.5\%$ & \phantom{0}$71.1\%$ \\ 
\midrule
\multirow{2}{*}{\gls*{plnet} (LR + SR)} & $3$ & \phantom{0}$17.0\%$ & \phantom{0}$45.5\%$ & \phantom{0}$69.0\%$ &  & \phantom{0}$17.7\%$ & \phantom{0}$45.7\%$ & \phantom{0}$69.3\%$ &  & \phantom{0}$18.3\%$ & \phantom{0}$47.3\%$ & \phantom{0}$70.7\%$  \\
& $5$ & \phantom{0}$18.3\%$ & \phantom{0}$47.0\%$ & \phantom{0}$72.5\%$ &  & \phantom{0}$20.7\%$ & \phantom{0}$49.1\%$ & \phantom{0}$73.7\%$ &  & \phantom{0}$22.7\%$ & \phantom{0}$53.4\%$ & \phantom{0}$76.9\%$ \\ 
\midrule
\multirow{2}{*}{\gls*{lcdnet} (LR + SR)} & $3$ & \phantom{0}$17.4\%$ & \phantom{0}$44.3\%$ & \phantom{0}$70.2\%$ & & \phantom{0}$18.5\%$ & \phantom{0}$45.4\%$ & \phantom{0}$70.5\%$ &  & \phantom{0}$19.3\%$ & \phantom{0}$48.4\%$ & \phantom{0}$72.2\%$ \\
\multicolumn{1}{l}{} & $5$ & \phantom{0}$18.5\%$ & \phantom{0}$46.2\%$ & \phantom{0}$72.4\%$ &  & \phantom{0}$21.7\%$ & \phantom{0}$49.7\%$ & \phantom{0}$73.9\%$ &  & \phantom{0}$23.5\%$ & \phantom{0}$55.3\%$ & \phantom{0}$78.2\%$ \\ 

\bottomrule
\end{tabular}
}
\end{table*}

\begin{table*}[!htb]
\centering
\caption{\major{Comparison of recognition rates achieved by combining the outputs of the \ocrchina model for multiple super-resolved images generated by PLNET and LCDNet (considering only \glspl*{lp} extracted from images with $1920\times1080$ pixels). As outlined in \cref{xp:fusioning}, three fusion strategies were evaluated: \acrfull*{hc}, \acrfull*{mv}, and \acrfull*{mvcp}.}}
\label{xp:quantitativeR35MJ1920x1080}
\resizebox{0.975\linewidth}{!}{
\begin{tabular}{lcccccccccccc} 
\toprule
\multirow{2}{*}{\begin{tabular}[c]{@{}c@{}}Test Images\\($\textbf{1920} \times \textbf{1080}$)\end{tabular}} & \multirow{2}{*}{\begin{tabular}[c]{@{}c@{}}\# Images \phantom{0}\\Majority Vote\end{tabular}} & \multicolumn{3}{c}{HC} &  & \multicolumn{3}{c}{MV} &  & \multicolumn{3}{c}{MVCP}  \\ 
\cmidrule{3-5} \cmidrule{7-9} \cmidrule{11-13}
&  & All & $\geq 6$ & $\geq 5$ &  & All & $\geq 6$ & $\geq 5$ &  & All & $\geq 6$ & $\geq 5$ \\ 

\midrule
\multirow{2}{*}{SR3 (LR + SR)} & $3$ & \phantom{0}$30.3\%$ & \phantom{0}$66.8\%$ & \phantom{0}$84.8\%$ &  & \phantom{0}$33.0\%$ & \phantom{0}$68.5\%$ & \phantom{0}$85.3\%$ &  & \phantom{0}$36.4\%$ & \phantom{0}$73.8\%$ & \phantom{0}$88.9\%$ \\ 
& $5$ & \phantom{0}$30.6\%$ & \phantom{0}$67.3\%$ & \phantom{0}$85.4\%$ &  & \phantom{0}$38.2\%$ & \phantom{0}$73.1\%$ & \phantom{0}$87.4\%$ &  & \phantom{0}$41.4\%$ & \phantom{0}$77.2\%$ & \phantom{0}$91.3\%$ \\
\midrule
\multicolumn{1}{l}{\multirow{2}{*}{\begin{tabular}[c]{@{}l@{}}LPSRGAN (LR + SR) \end{tabular}}} & $3$ & \phantom{0}$39.4\%$ & \phantom{0}$72.7\%$ & \phantom{0}$89.3\%$ &  & \phantom{0}$40.2\%$ & \phantom{0}$73.1\%$ & \phantom{0}$89.4\%$ &  & \phantom{0}$41.9\%$ & \phantom{0}$76.0\%$ & \phantom{0}$90.0\%$ \\ 
\multicolumn{1}{l}{} & $5$ & \phantom{0}$41.3\%$ & \phantom{0}$75.1\%$ & \phantom{0}$89.9\%$ &  & \phantom{0}$44.7\%$ & \phantom{0}$76.4\%$ & \phantom{0}$90.5\%$ &  & \phantom{0}$46.7\%$ & \phantom{0}$78.1\%$ & \phantom{0}$92.7\%$ \\ 
\midrule
\multirow{2}{*}{Real-ESRGAN (LR + SR)} & $3$ & \phantom{0}$34.8\%$ & \phantom{0}$71.5\%$ & \phantom{0}$88.4\%$ &  & \phantom{0}$35.7\%$ & \phantom{0}$72.7\%$ & \phantom{0}$88.5\%$ &  & \phantom{0}$37.5\%$ & \phantom{0}$75.1\%$ & \phantom{0}$90.8\%$ \\ 
\multicolumn{1}{l}{} & $5$ & \phantom{0}$35.8\%$ & \phantom{0}$72.8\%$ & \phantom{0}$89.6\%$ &  & \phantom{0}$39.8\%$ & \phantom{0}$76.3\%$ & \phantom{0}$90.3\%$ &  & \phantom{0}$41.6\%$ & \phantom{0}$78.0\%$ & \phantom{0}$92.1\%$ \\ 
\midrule
\multirow{2}{*}{\gls*{plnet} (LR + SR)} & $3$ & \phantom{0}$52.2\%$ & \phantom{0}$82.4\%$ & \phantom{0}$92.5\%$ &  & \phantom{0}$54.6\%$ & \phantom{0}$83.3\%$ & \phantom{0}$92.5\%$ &  & \phantom{0}$55.1\%$ & \phantom{0}$85.2\%$ & \phantom{0}$93.5\%$ \\ 
& $5$ & \phantom{0}$53.7\%$ & \phantom{0}$83.7\%$ & \phantom{0}$93.2\%$ &  & \phantom{0}$58.4\%$ & \phantom{0}$85.9\%$ & \phantom{0}$93.7\%$ &  & \phantom{0}$59.3\%$ & \phantom{0}$87.7\%$ & \phantom{0}$94.9\%$ \\
\midrule
\multirow{2}{*}{\gls*{lcdnet} (LR + SR)} & $3$ & \phantom{0}$52.3\%$ & \phantom{0}$83.0\%$ & \phantom{0}$93.6\%$ &  & \phantom{0}$55.1\%$ & \phantom{0}$83.8\%$ & \phantom{0}$93.7\%$ &  & \phantom{0}$56.4\%$ & \phantom{0}$85.5\%$ & \phantom{0}$94.5\%$  \\
\multicolumn{1}{l}{} & $5$ & \phantom{0}$53.1\%$ & \phantom{0}$83.6\%$ & \phantom{0}$93.8\%$ &  & \phantom{0}$59.8\%$ & \phantom{0}$86.7\%$ & \phantom{0}$94.5\%$ &  & \phantom{0}$61.4\%$ & \phantom{0}$89.0\%$ & \phantom{0}$95.8\%$  \\ 

\bottomrule
\end{tabular}
}
\end{table*}

\major{The sharp decline in recognition rates for \gls*{lr} images underscores the challenges of achieving accurate \gls*{lpr} when working with low-quality inputs.
While the \gls*{gplpr} model achieves $85.2\%$ recognition rate on \gls*{hr} images, its performance plunges to just $2.2\%$ on \gls*{lr} images, with only $20.1\%$ of \glspl*{lp} having at least $5$ correct characters. 
The application of \gls*{sr} methods, however, brings significant improvements.
Domain-specific models such as \gls*{lcdnet} and \gls*{plnet} achieve a $29.9\%$ recognition rate, outperforming general-purpose SR approaches like \acrshort*{sr3}~($18.4\%$) and \acrshort*{realesrgan}~($20.2\%$).
For partial recognition (at least five correct characters), \gls*{lcdnet} reaches $77.1$\%, demonstrating that even imperfect reconstructions can effectively reduce the search space in forensic scenarios.
Interestingly, \acrshort*{lpsrgan} --~despite being tailored for \glspl*{lp}~-- underperforms compared to \acrshort*{realesrgan}~($19.6$\% vs. $20.2$\%), indicating potential limitations in its degradation modeling or training methodology.
Overall, these findings highlight the promise of \gls*{sr} techniques in improving \gls*{lpr} robustness under real-world conditions. 
Nevertheless, the considerable gap between super-resolved~($29.9$\%) and \gls*{hr}~($85.2$\%) performance reveals the continued need for advancements in dealing with noise, compression, and other practical image~degradations.}

\major{\Cref{xp:quantitativeR35MJ} presents the recognition rates achieved by combining the \gls*{ocr} model's predictions for three and five super-resolved images using the strategies described in \cref{xp:fusioning}. 
For \gls*{lp}-specialized models, \gls*{plnet} achieves up to 40.9\% recognition rate (all characters correct) with \gls*{mvcp} and 5 images, while \gls*{lcdnet} outperforms slightly at 42.3\%, highlighting their robustness in reconstructing critical character details. 
Both models surpass general-purpose methods like \acrshort*{sr3} (28.3\%) and \acrshort*{realesrgan} (29.5\%). The \gls*{mvcp} strategy consistently delivers the highest gains, improving \gls*{plnet}'s accuracy by 5.0\% (3 to 5 images) and \gls*{lcdnet}'s by 4.5\% over \gls*{hc} and \gls*{mv}. 
Aggregating five images instead of three further boosts performance; for example, \gls*{lcdnet} achieves 86.9\% for cases with at least 5 correct characters (vs. 83.3\% with 3 images), nearing practical utility for surveillance systems. These results underscore the value of temporal fusion and domain-specific \gls*{sr} in overcoming real-world degradations, where partial matches ($\geq$~5 characters) remain critical for forensic tasks.}

\major{\Cref{xp:quantitativeR35MJ1280x960,xp:quantitativeR35MJ1920x1080} demonstrate the critical impact of image resolution on \gls*{lp} super-resolution and subsequent \gls*{lpr} performance. 
For lower-resolution sources ($1280 \times 960$~pixels), \gls*{plnet} and \gls*{lcdnet} achieve modest recognition rates of $22.7\%$ and $23.5\%$ (all characters correct, $5$ images + \gls*{mvcp}), respectively. In contrast, for high-resolution sources ($1920 \times 1080$~pixels), these models reach $59.3\%$ (\gls*{plnet}) and $61.4\%$ (\gls*{lcdnet}) under the same conditions --~a $2.6\times$ improvement~-- underscoring the importance of pixel density in preserving structural details like character edges and serifs.}

\major{The \gls*{mvcp} fusion strategy consistently outperforms \gls*{hc} and \gls*{mv} across resolutions. For $1280 \times 960$~images, \gls*{mvcp} boosts \gls*{lcdnet}'s accuracy by $+5.0\%$ ($18.5\% \rightarrow 23.5\%$) with $5$~images, while for $1920 \times 1080$~images, it improves results by $+7.7\%$ ($53.1\% \rightarrow 61.4\%$). Increasing the number of fused images from $3$ to $5$ further enhances performance: partial matches ($\geq5$~characters) rise from $78.2\%$ to $95.8\%$ for \gls*{lcdnet} in \gls*{hr} settings, demonstrating near-practical utility for surveillance systems.}

\major{Notably, even \gls*{lr} scenarios benefit significantly from fusion. 
For $1280 \times 960$~images, \gls*{mvcp} with 5~images achieves $71.6\%$ (\gls*{sr3}) to $78.2\%$ (\gls*{lcdnet}) for $\geq5$~correct characters, narrowing search spaces in forensic applications. This is critical for real-world deployments, where cost-effective cameras (e.g., $1280 \times 960$~sensors) dominate, and partial matches remain acceptable due to resolution constraints. The proposed dataset bridges this gap, enabling robust evaluation of \gls*{sr} methods across diverse operational~scenarios.}

\major{While recent \gls*{lp}-specific \gls*{sr} methods (e.g., \acrshort*{lpsrgan}~\citep{pan2024lpsrgan} and \gls*{plnet}~\citep{nascimento2023super,nascimento2024superCTD}) and general-purpose approaches (e.g., \acrshort*{realesrgan}~\citep{wang2021real} and SR3~\citep{saharia2023image}) have advanced super-resolution research, our experiments reveal their limitations in real-world scenarios. 
\acrshort*{lpsrgan}, despite its domain-specific design, achieves only \(19.6\%\) recognition accuracy on \dataset (\cref{xp:quantitativeR1MJ}), a drastic drop from the \(93.9\%\) it attains on synthetic benchmarks like LicensePlateDataset10K~\citep{pan2024lpsrgan}.
Similarly, \gls*{plnet}, SR3, and \gls*{lcdnet} --~which achieve \(49.8\%\), \(43.1\%\), and \(39.0\%\) on synthetic \rodosolalpr~\citep{nascimento2024enhancing} --~exhibit significantly lower accuracy on real-world \dataset (\(29.9\%\), \(18.4\%\), and \(29.9\%\), respectively; \cref{xp:quantitativeR1MJ}). 
This disparity underscores the limitations of synthetic degradation pipelines, which fail to replicate real-world noise patterns (e.g., motion blur, sensor noise, weather effects). 
\gls*{lcdnet}’s layout-aware architecture and \gls*{ocr}-guided training partially mitigate these challenges, achieving \(61.4\%\) accuracy with \gls*{mvcp} fusion (\cref{xp:quantitativeR35MJ}), but the gap between synthetic and real-world performance persists. 
These results emphasize the necessity of domain-specific architectures and real-world benchmarks like \dataset to advance robust \gls*{lpr} systems.}

\ifarxiv
    \begin{figure*}[!htb]

\vspace{0.7mm}

\centering

\resizebox{0.997\linewidth}{!}{
\begin{tabular}{
c|
c@{\hspace{0.5mm}}
c@{\hspace{0.5mm}}
c@{\hspace{0.5mm}}
c@{\hspace{0.5mm}}
c
c@{\hspace{0.5mm}}
c@{\hspace{0.5mm}}
c@{\hspace{0.5mm}}
c@{\hspace{0.5mm}}
c@{\hspace{0.5mm}}
c
}
\multirow{2}{*}{{\vspace{10mm}LR Inputs}} & 
\includegraphics[width=0.15\textwidth, height=0.06\textwidth]{4-Experiments/LCDnet-plate_1280x960/plate-000056/lr-001} & 
\includegraphics[width=0.15\textwidth, height=0.06\textwidth]{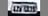} & 
\includegraphics[width=0.15\textwidth, height=0.06\textwidth]{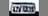} & 
\includegraphics[width=0.15\textwidth, height=0.06\textwidth]{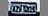} & 
\includegraphics[width=0.15\textwidth, height=0.06\textwidth]{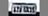} &  & 
\includegraphics[width=0.15\textwidth, height=0.06\textwidth]{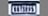} & 
\includegraphics[width=0.15\textwidth, height=0.06\textwidth]{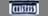} & 
\includegraphics[width=0.15\textwidth, height=0.06\textwidth]{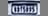} & 
\includegraphics[width=0.15\textwidth, height=0.06\textwidth]{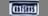} & 
\includegraphics[width=0.15\textwidth, height=0.06\textwidth]{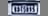} \\
& 
\textcolor{red}{LCI0D33} & 
\textcolor{blue}{A}\textcolor{red}{JX0D39} & 
\textcolor{blue}{A}\textcolor{red}{3V0336} & 
\textcolor{blue}{AZ}\textcolor{red}{Y0133} & 
\textcolor{red}{ATX0Q71} &  & 
\textcolor{blue}{B}\textcolor{red}{AJ0H01} & 
\textcolor{red}{AA}\textcolor{blue}{T}\textcolor{red}{8A91} & 
\textcolor{red}{AQ}\textcolor{blue}{T}\textcolor{red}{8991} & 
\textcolor{red}{AA}\textcolor{blue}{T}\textcolor{red}{1A91} & 
\textcolor{red}{AA}\textcolor{blue}{T}\textcolor{red}{4H11} \\
&  &  &  &  &  &  &  &  &  &  &  \\

\multirow{2}{*}{{\vspace{10mm}SR3}} & 
\includegraphics[width=0.15\textwidth, height=0.06\textwidth]{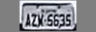} & 
\includegraphics[width=0.15\textwidth, height=0.06\textwidth]{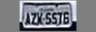} & 
\includegraphics[width=0.15\textwidth, height=0.06\textwidth]{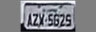} & 
\includegraphics[width=0.15\textwidth, height=0.06\textwidth]{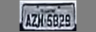} & 
\includegraphics[width=0.15\textwidth, height=0.06\textwidth]{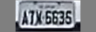} & & 

\includegraphics[width=0.15\textwidth, height=0.06\textwidth]{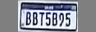} & 
\includegraphics[width=0.15\textwidth, height=0.06\textwidth]{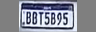} & 
\includegraphics[width=0.15\textwidth, height=0.06\textwidth]{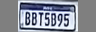} & 
\includegraphics[width=0.15\textwidth, height=0.06\textwidth]{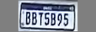} &  
\includegraphics[width=0.15\textwidth, height=0.06\textwidth]{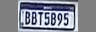} \\
& 
\textcolor{blue}{AZW5}\textcolor{red}{6}\textcolor{blue}{35} & 
\textcolor{blue}{AZ}\textcolor{red}{K}\textcolor{blue}{5}\textcolor{red}{576} & 
\textcolor{blue}{AZ}\textcolor{red}{V}\textcolor{blue}{5}\textcolor{red}{62}\textcolor{blue}{5} & 
\textcolor{blue}{AZW58}\textcolor{red}{29} & 
\textcolor{blue}{A}\textcolor{red}{TX60}\textcolor{blue}{35} &  & 

\textcolor{blue}{BBT5B95} & 
\textcolor{blue}{BBT5B95} & 
\textcolor{blue}{BBT5B95} & 
\textcolor{blue}{BBT5B95} & 
\textcolor{blue}{BBT5B95} \\ 

&  &  &  &  &  &  &  &  &  &  &  \\

\multirow{2}{*}{{\vspace{10mm}LPSRGAN}} & 
\includegraphics[width=0.15\textwidth, height=0.06\textwidth]{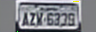} & 
\includegraphics[width=0.15\textwidth, height=0.06\textwidth]{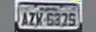} & 
\includegraphics[width=0.15\textwidth, height=0.06\textwidth]{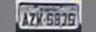} & 
\includegraphics[width=0.15\textwidth, height=0.06\textwidth]{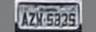} & 
\includegraphics[width=0.15\textwidth, height=0.06\textwidth]{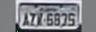} & & 

\includegraphics[width=0.15\textwidth, height=0.06\textwidth]{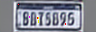} & 
\includegraphics[width=0.15\textwidth, height=0.06\textwidth]{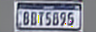} & 
\includegraphics[width=0.15\textwidth, height=0.06\textwidth]{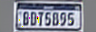} & 
\includegraphics[width=0.15\textwidth, height=0.06\textwidth]{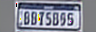} &  
\includegraphics[width=0.15\textwidth, height=0.06\textwidth]{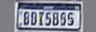} \\
& 
\textcolor{blue}{AZW}\textcolor{red}{63}\textcolor{blue}{3}\textcolor{red}{9} & 
\textcolor{blue}{AZW}\textcolor{red}{637}\textcolor{blue}{5} & 
\textcolor{blue}{AZW58}\textcolor{red}{79} & 
\textcolor{blue}{AZW5835} & 
\textcolor{blue}{AZW}\textcolor{red}{6}\textcolor{blue}{8}\textcolor{red}{7}\textcolor{blue}{5} &  & 

\textcolor{blue}{B}\textcolor{red}{D}\textcolor{blue}{T}\textcolor{blue}{5B95} & 
\textcolor{blue}{B}\textcolor{red}{D}\textcolor{blue}{T}\textcolor{blue}{5B95} & 
\textcolor{blue}{B}\textcolor{red}{D}\textcolor{blue}{T}\textcolor{blue}{5B95} & 
\textcolor{blue}{BBT5B95} & 
\textcolor{blue}{B}\textcolor{red}{D}\textcolor{blue}{T}\textcolor{blue}{5B}5\textcolor{blue}{5} \\ 

&  &  &  &  &  &  &  &  &  &  &  \\

\multirow{2}{*}{{\vspace{10mm}Real-ESRGAN}} & 
\includegraphics[width=0.15\textwidth, height=0.06\textwidth]{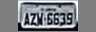} & 
\includegraphics[width=0.15\textwidth, height=0.06\textwidth]{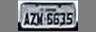} & 
\includegraphics[width=0.15\textwidth, height=0.06\textwidth]{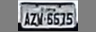} & 
\includegraphics[width=0.15\textwidth, height=0.06\textwidth]{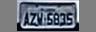} & 
\includegraphics[width=0.15\textwidth, height=0.06\textwidth]{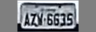} & & 

\includegraphics[width=0.15\textwidth, height=0.06\textwidth]{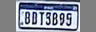} & 
\includegraphics[width=0.15\textwidth, height=0.06\textwidth]{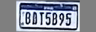} & 
\includegraphics[width=0.15\textwidth, height=0.06\textwidth]{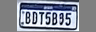} & 
\includegraphics[width=0.15\textwidth, height=0.06\textwidth]{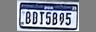} &  
\includegraphics[width=0.15\textwidth, height=0.06\textwidth]{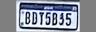} \\
& 
\textcolor{blue}{AZW}\textcolor{red}{66}\textcolor{blue}{3}\textcolor{red}{9} & 
\textcolor{blue}{AZW}\textcolor{blue}{5}\textcolor{red}{6}\textcolor{blue}{35} & 
\textcolor{blue}{AZW}\textcolor{red}{657}\textcolor{blue}{5} & 
\textcolor{blue}{AZ}\textcolor{red}{N}\textcolor{blue}{5835} & 
\textcolor{blue}{AZ}\textcolor{red}{266}\textcolor{blue}{35} &  & 

\textcolor{blue}{B}\textcolor{red}{D}\textcolor{blue}{T}\textcolor{red}{3}\textcolor{blue}{B95} & 
\textcolor{blue}{B}\textcolor{red}{A}\textcolor{blue}{T}\textcolor{blue}{5B95} & 
\textcolor{blue}{B}\textcolor{red}{D}\textcolor{blue}{T}\textcolor{blue}{5B95} & 
\textcolor{blue}{B}\textcolor{red}{D}\textcolor{blue}{T}\textcolor{blue}{5B95} & 
\textcolor{blue}{B}\textcolor{red}{D}\textcolor{blue}{T}\textcolor{blue}{5B}\textcolor{red}{3}\textcolor{blue}{5} \\ 

&  &  &  &  &  &  &  &  &  &  &  \\

\multirow{2}{*}{{\vspace{10mm}\gls*{plnet}}} & 
\includegraphics[width=0.15\textwidth, height=0.06\textwidth]{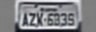} & 
\includegraphics[width=0.15\textwidth, height=0.06\textwidth]{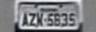} & 
\includegraphics[width=0.15\textwidth, height=0.06\textwidth]{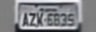} & 
\includegraphics[width=0.15\textwidth, height=0.06\textwidth]{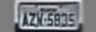} &  
\includegraphics[width=0.15\textwidth, height=0.06\textwidth]{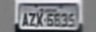} &  & 

\includegraphics[width=0.15\textwidth, height=0.06\textwidth]{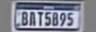} & 
\includegraphics[width=0.15\textwidth, height=0.06\textwidth]{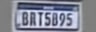} & 
\includegraphics[width=0.15\textwidth, height=0.06\textwidth]{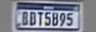} & 
\includegraphics[width=0.15\textwidth, height=0.06\textwidth]{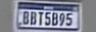} &  
\includegraphics[width=0.15\textwidth, height=0.06\textwidth]{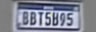} \\
&
\textcolor{blue}{AZ}\textcolor{red}{K63}\textcolor{blue}{35} &
\textcolor{blue}{AZ}\textcolor{red}{K63}\textcolor{blue}{35} &
\textcolor{blue}{AZ}\textcolor{red}{K6I}\textcolor{blue}{35} &
\textcolor{blue}{AZW5835} &
\textcolor{blue}{AZ}\textcolor{red}{X66}\textcolor{blue}{35} &  & 

\textcolor{blue}{B}\textcolor{red}{A}\textcolor{blue}{T5B95} & 
\textcolor{blue}{B}\textcolor{red}{R}\textcolor{blue}{T5B95} & 
\textcolor{blue}{B}\textcolor{red}{D}\textcolor{blue}{T5B95} & 
\textcolor{blue}{BBT5B95} & 
\textcolor{blue}{BBT5}\textcolor{red}{H}\textcolor{blue}{95} \\

&  &  &  &  &  &  &  &  &  &  &  \\

\multirow{2}{*}{{\vspace{10mm}\gls*{lcdnet}}} & 
\includegraphics[width=0.15\textwidth, height=0.06\textwidth]{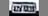} & 
\includegraphics[width=0.15\textwidth, height=0.06\textwidth]{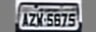} & 
\includegraphics[width=0.15\textwidth, height=0.06\textwidth]{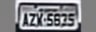} & 
\includegraphics[width=0.15\textwidth, height=0.06\textwidth]{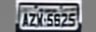} & 
\includegraphics[width=0.15\textwidth, height=0.06\textwidth]{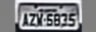} & & 

\includegraphics[width=0.15\textwidth, height=0.06\textwidth]{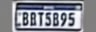} & 
\includegraphics[width=0.15\textwidth, height=0.06\textwidth]{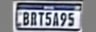} & 
\includegraphics[width=0.15\textwidth, height=0.06\textwidth]{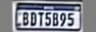} & 
\includegraphics[width=0.15\textwidth, height=0.06\textwidth]{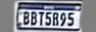} &  
\includegraphics[width=0.15\textwidth, height=0.06\textwidth]{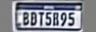} \\
& 
\textcolor{blue}{AZW5}\textcolor{red}{639} & 
\textcolor{blue}{AZW58}\textcolor{red}{7}\textcolor{blue}{5}& 
\textcolor{blue}{AZ}\textcolor{red}{K}\textcolor{blue}{5835}& 
\textcolor{blue}{AZW58}\textcolor{red}{2}\textcolor{blue}{5} & 
\textcolor{blue}{AZW5}\textcolor{red}{1}\textcolor{blue}{35} &  & 

\textcolor{blue}{BBT5B95} & 
\textcolor{blue}{BAT5}\textcolor{red}{A}\textcolor{blue}{95} & 
\textcolor{blue}{B}\textcolor{red}{D}\textcolor{blue}{T5B95} & 
\textcolor{blue}{BBT5B95} & 
\textcolor{blue}{BBT5B95} \\ 

\hdashline

&  &  &  &  &  &  &  &  &  &  &  \\

\multirow{2}{*}{{\vspace{10mm}HR}} &   
&  & \includegraphics[width=0.15\textwidth, height=0.06\textwidth]{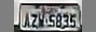} &
&   & &   
&  & \includegraphics[width=0.15\textwidth, height=0.06\textwidth]{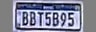} &  &  \\
&  &  & \textcolor{green}{AZW5835} &  &  &  &  &  & \textcolor{green}{BBT5B95} &  &  \\
\end{tabular}
}

\vspace{0.3mm}
\caption{Recognition results for \glspl*{lp} cropped from images with a resolution of $1280\times960$ pixels. The top row shows the predictions made by \ocrchina~\citep{liu2024irregular} on the original \gls*{lr} images, while the two subsequent rows present the predictions obtained from super-resolved images generated by \gls*{plnet}~\citep{nascimento2023super,nascimento2024superCTD} and \gls*{lcdnet}~\citep{nascimento2024enhancing}. 
Below each image, the predicted characters are displayed, with correct characters highlighted in blue and incorrect characters in red. The ground truth is indicated in green.}
\label{fig:qualitative_rsults_rr1}
\end{figure*}

    \begin{figure*}[!htb]

\vspace{0.7mm}

\centering

\resizebox{0.997\linewidth}{!}{
\begin{tabular}{
c|
c@{\hspace{0.5mm}}
c@{\hspace{0.5mm}}
c@{\hspace{0.5mm}}
c@{\hspace{0.5mm}}
c
c@{\hspace{0.5mm}}
c@{\hspace{0.5mm}}
c@{\hspace{0.5mm}}
c@{\hspace{0.5mm}}
c@{\hspace{0.5mm}}
c
}
\multirow{2}{*}{{\vspace{10mm}LR Inputs}} & 
\includegraphics[width=0.15\textwidth, height=0.06\textwidth]{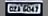} & 
\includegraphics[width=0.15\textwidth, height=0.06\textwidth]{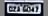} & 
\includegraphics[width=0.15\textwidth, height=0.06\textwidth]{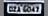} & 
\includegraphics[width=0.15\textwidth, height=0.06\textwidth]{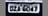} & 
\includegraphics[width=0.15\textwidth, height=0.06\textwidth]{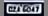} &  & 
\includegraphics[width=0.15\textwidth, height=0.06\textwidth]{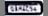} & 
\includegraphics[width=0.15\textwidth, height=0.06\textwidth]{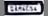} & 
\includegraphics[width=0.15\textwidth, height=0.06\textwidth]{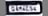} & 
\includegraphics[width=0.15\textwidth, height=0.06\textwidth]{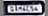} & 
\includegraphics[width=0.15\textwidth, height=0.06\textwidth]{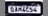} \\
& 
\textcolor{red}{OPI9C}\textcolor{blue}{47} & 
\textcolor{red}{C0A4A}\textcolor{blue}{47} & 
\textcolor{red}{OPI}\textcolor{blue}{6}\textcolor{red}{8}\textcolor{blue}{47} & 
\textcolor{red}{OZI}\textcolor{blue}{6}\textcolor{red}{C}\textcolor{blue}{47} & 
\textcolor{red}{CBI}\textcolor{blue}{6}\textcolor{red}{9}\textcolor{blue}{47} &  & 
\textcolor{blue}{Q}\textcolor{red}{I}\textcolor{blue}{M4C}\textcolor{red}{1}\textcolor{blue}{4} & 
\textcolor{red}{E}\textcolor{blue}{AM4}\textcolor{red}{E3}\textcolor{blue}{4} & 
\textcolor{red}{7}\textcolor{blue}{AM}\textcolor{red}{6}\textcolor{blue}{C}\textcolor{red}{8}\textcolor{blue}{4} & 
\textcolor{red}{PV}\textcolor{blue}{M4}\textcolor{red}{E1}\textcolor{blue}{4} & 
\textcolor{red}{OJ}\textcolor{blue}{M}\textcolor{red}{2Z5}\textcolor{blue}{4} \\
&  &  &  &  &  &  &  &  &  &  &  \\

\multirow{2}{*}{{\vspace{10mm}SR3}} & 
\includegraphics[width=0.15\textwidth, height=0.06\textwidth]{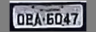} & 
\includegraphics[width=0.15\textwidth, height=0.06\textwidth]{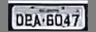} & 
\includegraphics[width=0.15\textwidth, height=0.06\textwidth]{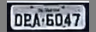} & 
\includegraphics[width=0.15\textwidth, height=0.06\textwidth]{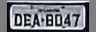} & 
\includegraphics[width=0.15\textwidth, height=0.06\textwidth]{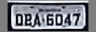} & & 

\includegraphics[width=0.15\textwidth, height=0.06\textwidth]{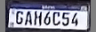} & 
\includegraphics[width=0.15\textwidth, height=0.06\textwidth]{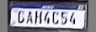} & 
\includegraphics[width=0.15\textwidth, height=0.06\textwidth]{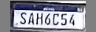} & 
\includegraphics[width=0.15\textwidth, height=0.06\textwidth]{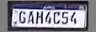} &  
\includegraphics[width=0.15\textwidth, height=0.06\textwidth]{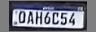} \\
& 
\textcolor{red}{OB}\textcolor{blue}{A6}\textcolor{red}{6}\textcolor{blue}{47} &     
\textcolor{blue}{D}\textcolor{red}{B}\textcolor{blue}{A6}\textcolor{red}{6}\textcolor{blue}{47} & 
\textcolor{blue}{D}\textcolor{red}{4}\textcolor{blue}{A6}\textcolor{red}{6}\textcolor{blue}{47} & 
\textcolor{blue}{D}\textcolor{red}{E}\textcolor{blue}{A}\textcolor{red}{8}\textcolor{blue}{047} & 
\textcolor{red}{QB}\textcolor{blue}{A6}\textcolor{red}{6}\textcolor{blue}{47} &  & 
 \textcolor{red}{G}\textcolor{blue}{A}\textcolor{red}{H6}\textcolor{blue}{C54} & 
 \textcolor{red}{G}\textcolor{blue}{A}\textcolor{red}{H}\textcolor{blue}{4C54} & 
 \textcolor{red}{S}\textcolor{blue}{A}\textcolor{red}{H6}\textcolor{blue}{C54} & 
 \textcolor{red}{G}\textcolor{blue}{A}\textcolor{red}{H6}\textcolor{blue}{C54} & 
 \textcolor{blue}{QA}\textcolor{red}{H6}\textcolor{blue}{C54} \\  

&  &  &  &  &  &  &  &  &  &  &  \\

\multirow{2}{*}{{\vspace{10mm}LPSRGAN}} & 
\includegraphics[width=0.15\textwidth, height=0.06\textwidth]{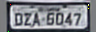} & 
\includegraphics[width=0.15\textwidth, height=0.06\textwidth]{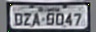} & 
\includegraphics[width=0.15\textwidth, height=0.06\textwidth]{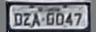} & 
\includegraphics[width=0.15\textwidth, height=0.06\textwidth]{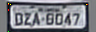} & 
\includegraphics[width=0.15\textwidth, height=0.06\textwidth]{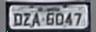} & & 

\includegraphics[width=0.15\textwidth, height=0.06\textwidth]{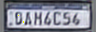} & 
\includegraphics[width=0.15\textwidth, height=0.06\textwidth]{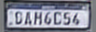} & 
\includegraphics[width=0.15\textwidth, height=0.06\textwidth]{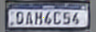} & 
\includegraphics[width=0.15\textwidth, height=0.06\textwidth]{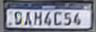} &  
\includegraphics[width=0.15\textwidth, height=0.06\textwidth]{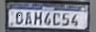} \\
& 
\textcolor{blue}{DZA6047} & 
\textcolor{blue}{DZA6047} & 
\textcolor{blue}{DZA6047} & 
\textcolor{red}{O}\textcolor{blue}{ZA}\textcolor{red}{8}\textcolor{blue}{047} & 
\textcolor{blue}{DZA6047} &  & 

\textcolor{red}{O}\textcolor{blue}{AM}\textcolor{red}{6}\textcolor{blue}{C54} & 
\textcolor{red}{O}\textcolor{blue}{AM}\textcolor{red}{6}\textcolor{blue}{C54} & 
\textcolor{red}{O}\textcolor{blue}{A}\textcolor{red}{H6}\textcolor{blue}{C54} & 
\textcolor{blue}{QAM4C54} & 
\textcolor{red}{O}\textcolor{blue}{AM}\textcolor{red}{6}\textcolor{blue}{C54} \\ 

&  &  &  &  &  &  &  &  &  &  &  \\

\multirow{2}{*}{{\vspace{10mm}Real-ESRGAN}} & 
\includegraphics[width=0.15\textwidth, height=0.06\textwidth]{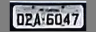} & 
\includegraphics[width=0.15\textwidth, height=0.06\textwidth]{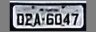} & 
\includegraphics[width=0.15\textwidth, height=0.06\textwidth]{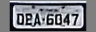} & 
\includegraphics[width=0.15\textwidth, height=0.06\textwidth]{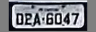} & 
\includegraphics[width=0.15\textwidth, height=0.06\textwidth]{4-Experiments/Real-ESRGAN-1920x1080/plate-000253/sr_004.png} & & 

\includegraphics[width=0.15\textwidth, height=0.06\textwidth]{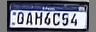} & 
\includegraphics[width=0.15\textwidth, height=0.06\textwidth]{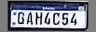} & 
\includegraphics[width=0.15\textwidth, height=0.06\textwidth]{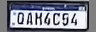} & 
\includegraphics[width=0.15\textwidth, height=0.06\textwidth]{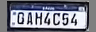} &  
\includegraphics[width=0.15\textwidth, height=0.06\textwidth]{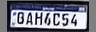} \\
& 
\textcolor{blue}{DZA6}\textcolor{red}{6}\textcolor{blue}{47} & 
\textcolor{blue}{DZA6}\textcolor{red}{6}\textcolor{blue}{47} & 
\textcolor{blue}{DBA6}\textcolor{red}{6}\textcolor{blue}{47} & 
\textcolor{blue}{DZA6}\textcolor{red}{6}\textcolor{blue}{47} & 
\textcolor{red}{O}\textcolor{blue}{ZA6}\textcolor{red}{6}\textcolor{blue}{47} &  & 
\textcolor{blue}{QA}\textcolor{red}{H6}\textcolor{blue}{C54} & 
\textcolor{blue}{QA}\textcolor{red}{H6}\textcolor{blue}{C54} & 
\textcolor{blue}{QA}\textcolor{red}{M}\textcolor{blue}{4C54} & 
\textcolor{red}{O}\textcolor{blue}{AH4C54} & 
\textcolor{red}{O}\textcolor{blue}{AH}\textcolor{red}{6}\textcolor{blue}{C54} \\  

&  &  &  &  &  &  &  &  &  &  &  \\

\multirow{2}{*}{{\vspace{10mm}\gls*{plnet}}} & 
\includegraphics[width=0.15\textwidth, height=0.06\textwidth]{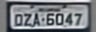} & 
\includegraphics[width=0.15\textwidth, height=0.06\textwidth]{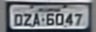} & 
\includegraphics[width=0.15\textwidth, height=0.06\textwidth]{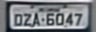} & 
\includegraphics[width=0.15\textwidth, height=0.06\textwidth]{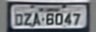} &  
\includegraphics[width=0.15\textwidth, height=0.06\textwidth]{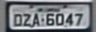} &  & 

\includegraphics[width=0.15\textwidth, height=0.06\textwidth]{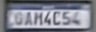} & 
\includegraphics[width=0.15\textwidth, height=0.06\textwidth]{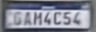} & 
\includegraphics[width=0.15\textwidth, height=0.06\textwidth]{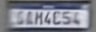} & 
\includegraphics[width=0.15\textwidth, height=0.06\textwidth]{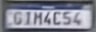} &  
\includegraphics[width=0.15\textwidth, height=0.06\textwidth]{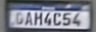} \\
& 
\textcolor{red}{O}\textcolor{blue}{ZA}\textcolor{blue}{6047} & 
\textcolor{red}{O}\textcolor{blue}{ZA}\textcolor{blue}{6047} & 
\textcolor{red}{O}\textcolor{blue}{ZA}\textcolor{blue}{6047} & 
\textcolor{red}{O}\textcolor{blue}{ZA}\textcolor{red}{8}\textcolor{blue}{047} & 
\textcolor{red}{O}\textcolor{blue}{ZA6047}  &  & 

\textcolor{red}{D}\textcolor{blue}{AM4C54} & 
\textcolor{red}{G}\textcolor{blue}{AM4C54} & 
\textcolor{red}{S}\textcolor{blue}{AM4C54} & 
\textcolor{red}{GI}\textcolor{blue}{M4C54} & 
\textcolor{blue}{QA}\textcolor{red}{H}\textcolor{blue}{4C54}  \\
&  &  &  &  &  &  &  &  &  &  &  \\
\multirow{2}{*}{{\vspace{10mm}\gls*{lcdnet}}} & 
\includegraphics[width=0.15\textwidth, height=0.06\textwidth]{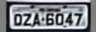} & 
\includegraphics[width=0.15\textwidth, height=0.06\textwidth]{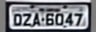} & 
\includegraphics[width=0.15\textwidth, height=0.06\textwidth]{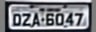} & 
\includegraphics[width=0.15\textwidth, height=0.06\textwidth]{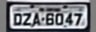} & 
\includegraphics[width=0.15\textwidth, height=0.06\textwidth]{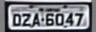} & & 

\includegraphics[width=0.15\textwidth, height=0.06\textwidth]{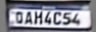} & 
\includegraphics[width=0.15\textwidth, height=0.06\textwidth]{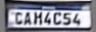} & 
\includegraphics[width=0.15\textwidth, height=0.06\textwidth]{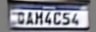} & 
\includegraphics[width=0.15\textwidth, height=0.06\textwidth]{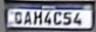} &  
\includegraphics[width=0.15\textwidth, height=0.06\textwidth]{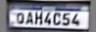} \\
& 
\textcolor{red}{Q}\textcolor{blue}{ZA6047} & 
\textcolor{red}{O}\textcolor{blue}{ZA6047}& 
\textcolor{red}{O}\textcolor{blue}{ZA6047} & 
\textcolor{red}{Q}\textcolor{blue}{ZA}\textcolor{red}{8}\textcolor{blue}{047} & 
\textcolor{red}{O}\textcolor{blue}{ZA6047} &  & 
\textcolor{blue}{QAM4C54} & 
\textcolor{red}{C}\textcolor{blue}{AM4C54} & 
\textcolor{blue}{QAM4C54} & 
\textcolor{blue}{QA}\textcolor{red}{H}\textcolor{blue}{4C54} & 
\textcolor{blue}{QA}\textcolor{red}{H}\textcolor{blue}{4C54} \\  

\hdashline

&  &  &  &  &  &  &  &  &  &  &  \\

\multirow{2}{*}{{\vspace{10mm}HR}} &   
&  & \includegraphics[width=0.15\textwidth, height=0.06\textwidth]{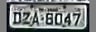} &
&   & &   
&  & \includegraphics[width=0.15\textwidth, height=0.06\textwidth]{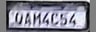} &  &  \\
&  &  & \textcolor{green}{DZA6047} &  &  &  &  &  & \textcolor{green}{QAM4C54} &  &  \\
\end{tabular}
}

\vspace{0.3mm}
\caption{Recognition results for \glspl*{lp} cropped from images with a resolution of $1920\times1080$ pixels. The top row shows the predictions made by \ocrchina~\citep{liu2024irregular} on the original \gls*{lr} images, while the two subsequent rows present the predictions obtained from super-resolved images generated by \gls*{plnet}~\citep{nascimento2023super,nascimento2024superCTD} and \gls*{lcdnet}~\citep{nascimento2024enhancing}. 
Below each image, the predicted characters are displayed, with correct characters highlighted in blue and incorrect characters in red. The ground truth is indicated in green.}
\label{fig:qualitative_rsults_rr2}
\end{figure*}

\else
    \begin{figure*}[!htb]

\vspace{0.7mm}

\centering

\resizebox{0.997\linewidth}{!}{
\begin{tabular}{
c|
c@{\hspace{0.5mm}}
c@{\hspace{0.5mm}}
c@{\hspace{0.5mm}}
c@{\hspace{0.5mm}}
c
c@{\hspace{0.5mm}}
c@{\hspace{0.5mm}}
c@{\hspace{0.5mm}}
c@{\hspace{0.5mm}}
c@{\hspace{0.5mm}}
c
}
\multirow{2}{*}{\addvbuffer[-3.5mm 8mm]{LR Inputs}} & 
\includegraphics[width=0.15\textwidth, height=0.06\textwidth]{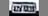} & 
\includegraphics[width=0.15\textwidth, height=0.06\textwidth]{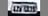} & 
\includegraphics[width=0.15\textwidth, height=0.06\textwidth]{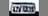} & 
\includegraphics[width=0.15\textwidth, height=0.06\textwidth]{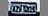} & 
\includegraphics[width=0.15\textwidth, height=0.06\textwidth]{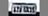} &  & 
\includegraphics[width=0.15\textwidth, height=0.06\textwidth]{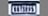} & 
\includegraphics[width=0.15\textwidth, height=0.06\textwidth]{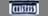} & 
\includegraphics[width=0.15\textwidth, height=0.06\textwidth]{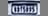} & 
\includegraphics[width=0.15\textwidth, height=0.06\textwidth]{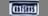} & 
\includegraphics[width=0.15\textwidth, height=0.06\textwidth]{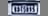} \\
& 
\textcolor{red}{LCI0D33} & 
\textcolor{blue}{A}\textcolor{red}{JX0D39} & 
\textcolor{blue}{A}\textcolor{red}{3V0336} & 
\textcolor{blue}{AZ}\textcolor{red}{Y0133} & 
\textcolor{red}{ATX0Q71} &  & 
\textcolor{blue}{B}\textcolor{red}{AJ0H01} & 
\textcolor{red}{AA}\textcolor{blue}{T}\textcolor{red}{8A91} & 
\textcolor{red}{AQ}\textcolor{blue}{T}\textcolor{red}{8991} & 
\textcolor{red}{AA}\textcolor{blue}{T}\textcolor{red}{1A91} & 
\textcolor{red}{AA}\textcolor{blue}{T}\textcolor{red}{4H11} \\
&  &  &  &  &  &  &  &  &  &  &  \\

\multirow{2}{*}{\addvbuffer[-3.5mm 8mm]{SR3}} & 
\includegraphics[width=0.15\textwidth, height=0.06\textwidth]{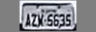} & 
\includegraphics[width=0.15\textwidth, height=0.06\textwidth]{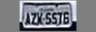} & 
\includegraphics[width=0.15\textwidth, height=0.06\textwidth]{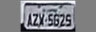} & 
\includegraphics[width=0.15\textwidth, height=0.06\textwidth]{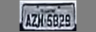} & 
\includegraphics[width=0.15\textwidth, height=0.06\textwidth]{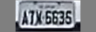} & & 

\includegraphics[width=0.15\textwidth, height=0.06\textwidth]{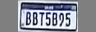} & 
\includegraphics[width=0.15\textwidth, height=0.06\textwidth]{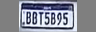} & 
\includegraphics[width=0.15\textwidth, height=0.06\textwidth]{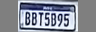} & 
\includegraphics[width=0.15\textwidth, height=0.06\textwidth]{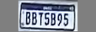} &  
\includegraphics[width=0.15\textwidth, height=0.06\textwidth]{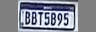} \\
& 
\textcolor{blue}{AZW5}\textcolor{red}{6}\textcolor{blue}{35} & 
\textcolor{blue}{AZ}\textcolor{red}{K}\textcolor{blue}{5}\textcolor{red}{576} & 
\textcolor{blue}{AZ}\textcolor{red}{V}\textcolor{blue}{5}\textcolor{red}{62}\textcolor{blue}{5} & 
\textcolor{blue}{AZW58}\textcolor{red}{29} & 
\textcolor{blue}{A}\textcolor{red}{TX60}\textcolor{blue}{35} &  & 

\textcolor{blue}{BBT5B95} & 
\textcolor{blue}{BBT5B95} & 
\textcolor{blue}{BBT5B95} & 
\textcolor{blue}{BBT5B95} & 
\textcolor{blue}{BBT5B95} \\ 

&  &  &  &  &  &  &  &  &  &  &  \\

\multirow{2}{*}{\addvbuffer[-3.5mm 8mm]{LPSRGAN}} & 
\includegraphics[width=0.15\textwidth, height=0.06\textwidth]{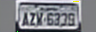} & 
\includegraphics[width=0.15\textwidth, height=0.06\textwidth]{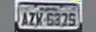} & 
\includegraphics[width=0.15\textwidth, height=0.06\textwidth]{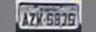} & 
\includegraphics[width=0.15\textwidth, height=0.06\textwidth]{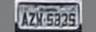} & 
\includegraphics[width=0.15\textwidth, height=0.06\textwidth]{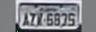} & & 

\includegraphics[width=0.15\textwidth, height=0.06\textwidth]{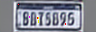} & 
\includegraphics[width=0.15\textwidth, height=0.06\textwidth]{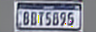} & 
\includegraphics[width=0.15\textwidth, height=0.06\textwidth]{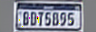} & 
\includegraphics[width=0.15\textwidth, height=0.06\textwidth]{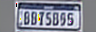} &  
\includegraphics[width=0.15\textwidth, height=0.06\textwidth]{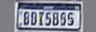} \\
& 
\textcolor{blue}{AZW}\textcolor{red}{63}\textcolor{blue}{3}\textcolor{red}{9} & 
\textcolor{blue}{AZW}\textcolor{red}{637}\textcolor{blue}{5} & 
\textcolor{blue}{AZW58}\textcolor{red}{79} & 
\textcolor{blue}{AZW5835} & 
\textcolor{blue}{AZW}\textcolor{red}{6}\textcolor{blue}{8}\textcolor{red}{7}\textcolor{blue}{5} &  & 

\textcolor{blue}{B}\textcolor{red}{D}\textcolor{blue}{T}\textcolor{blue}{5B95} & 
\textcolor{blue}{B}\textcolor{red}{D}\textcolor{blue}{T}\textcolor{blue}{5B95} & 
\textcolor{blue}{B}\textcolor{red}{D}\textcolor{blue}{T}\textcolor{blue}{5B95} & 
\textcolor{blue}{BBT5B95} & 
\textcolor{blue}{B}\textcolor{red}{D}\textcolor{blue}{T}\textcolor{blue}{5B}5\textcolor{blue}{5} \\ 

&  &  &  &  &  &  &  &  &  &  &  \\

\multirow{2}{*}{\addvbuffer[-3.5mm 8mm]{Real-ESRGAN}} & 
\includegraphics[width=0.15\textwidth, height=0.06\textwidth]{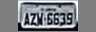} & 
\includegraphics[width=0.15\textwidth, height=0.06\textwidth]{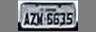} & 
\includegraphics[width=0.15\textwidth, height=0.06\textwidth]{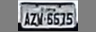} & 
\includegraphics[width=0.15\textwidth, height=0.06\textwidth]{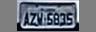} & 
\includegraphics[width=0.15\textwidth, height=0.06\textwidth]{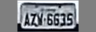} & & 

\includegraphics[width=0.15\textwidth, height=0.06\textwidth]{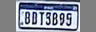} & 
\includegraphics[width=0.15\textwidth, height=0.06\textwidth]{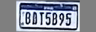} & 
\includegraphics[width=0.15\textwidth, height=0.06\textwidth]{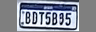} & 
\includegraphics[width=0.15\textwidth, height=0.06\textwidth]{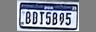} &  
\includegraphics[width=0.15\textwidth, height=0.06\textwidth]{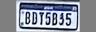} \\
& 
\textcolor{blue}{AZW}\textcolor{red}{66}\textcolor{blue}{3}\textcolor{red}{9} & 
\textcolor{blue}{AZW}\textcolor{blue}{5}\textcolor{red}{6}\textcolor{blue}{35} & 
\textcolor{blue}{AZW}\textcolor{red}{657}\textcolor{blue}{5} & 
\textcolor{blue}{AZ}\textcolor{red}{N}\textcolor{blue}{5835} & 
\textcolor{blue}{AZ}\textcolor{red}{266}\textcolor{blue}{35} &  & 

\textcolor{blue}{B}\textcolor{red}{D}\textcolor{blue}{T}\textcolor{red}{3}\textcolor{blue}{B95} & 
\textcolor{blue}{B}\textcolor{red}{A}\textcolor{blue}{T}\textcolor{blue}{5B95} & 
\textcolor{blue}{B}\textcolor{red}{D}\textcolor{blue}{T}\textcolor{blue}{5B95} & 
\textcolor{blue}{B}\textcolor{red}{D}\textcolor{blue}{T}\textcolor{blue}{5B95} & 
\textcolor{blue}{B}\textcolor{red}{D}\textcolor{blue}{T}\textcolor{blue}{5B}\textcolor{red}{3}\textcolor{blue}{5} \\ 

&  &  &  &  &  &  &  &  &  &  &  \\

\multirow{2}{*}{\addvbuffer[-3.5mm 8mm]{\gls*{plnet}}} & 
\includegraphics[width=0.15\textwidth, height=0.06\textwidth]{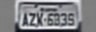} & 
\includegraphics[width=0.15\textwidth, height=0.06\textwidth]{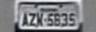} & 
\includegraphics[width=0.15\textwidth, height=0.06\textwidth]{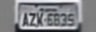} & 
\includegraphics[width=0.15\textwidth, height=0.06\textwidth]{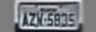} &  
\includegraphics[width=0.15\textwidth, height=0.06\textwidth]{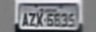} &  & 

\includegraphics[width=0.15\textwidth, height=0.06\textwidth]{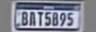} & 
\includegraphics[width=0.15\textwidth, height=0.06\textwidth]{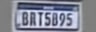} & 
\includegraphics[width=0.15\textwidth, height=0.06\textwidth]{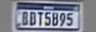} & 
\includegraphics[width=0.15\textwidth, height=0.06\textwidth]{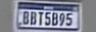} &  
\includegraphics[width=0.15\textwidth, height=0.06\textwidth]{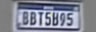} \\
&
\textcolor{blue}{AZ}\textcolor{red}{K63}\textcolor{blue}{35} &
\textcolor{blue}{AZ}\textcolor{red}{K63}\textcolor{blue}{35} &
\textcolor{blue}{AZ}\textcolor{red}{K6I}\textcolor{blue}{35} &
\textcolor{blue}{AZW5835} &
\textcolor{blue}{AZ}\textcolor{red}{X66}\textcolor{blue}{35} &  & 

\textcolor{blue}{B}\textcolor{red}{A}\textcolor{blue}{T5B95} & 
\textcolor{blue}{B}\textcolor{red}{R}\textcolor{blue}{T5B95} & 
\textcolor{blue}{B}\textcolor{red}{D}\textcolor{blue}{T5B95} & 
\textcolor{blue}{BBT5B95} & 
\textcolor{blue}{BBT5}\textcolor{red}{H}\textcolor{blue}{95} \\

&  &  &  &  &  &  &  &  &  &  &  \\

\multirow{2}{*}{\addvbuffer[-3.5mm 8mm]{\gls*{lcdnet}}} & 
\includegraphics[width=0.15\textwidth, height=0.06\textwidth]{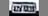} & 
\includegraphics[width=0.15\textwidth, height=0.06\textwidth]{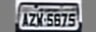} & 
\includegraphics[width=0.15\textwidth, height=0.06\textwidth]{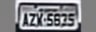} & 
\includegraphics[width=0.15\textwidth, height=0.06\textwidth]{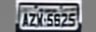} & 
\includegraphics[width=0.15\textwidth, height=0.06\textwidth]{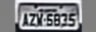} & & 

\includegraphics[width=0.15\textwidth, height=0.06\textwidth]{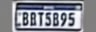} & 
\includegraphics[width=0.15\textwidth, height=0.06\textwidth]{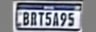} & 
\includegraphics[width=0.15\textwidth, height=0.06\textwidth]{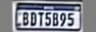} & 
\includegraphics[width=0.15\textwidth, height=0.06\textwidth]{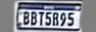} &  
\includegraphics[width=0.15\textwidth, height=0.06\textwidth]{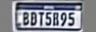} \\
& 
\textcolor{blue}{AZW5}\textcolor{red}{639} & 
\textcolor{blue}{AZW58}\textcolor{red}{7}\textcolor{blue}{5}& 
\textcolor{blue}{AZ}\textcolor{red}{K}\textcolor{blue}{5835}& 
\textcolor{blue}{AZW58}\textcolor{red}{2}\textcolor{blue}{5} & 
\textcolor{blue}{AZW5}\textcolor{red}{1}\textcolor{blue}{35} &  & 

\textcolor{blue}{BBT5B95} & 
\textcolor{blue}{BAT5}\textcolor{red}{A}\textcolor{blue}{95} & 
\textcolor{blue}{B}\textcolor{red}{D}\textcolor{blue}{T5B95} & 
\textcolor{blue}{BBT5B95} & 
\textcolor{blue}{BBT5B95} \\ 

\hdashline

&  &  &  &  &  &  &  &  &  &  &  \\

\multirow{2}{*}{\addvbuffer[-3.5mm 8mm]{HR}} &   
&  & \includegraphics[width=0.15\textwidth, height=0.06\textwidth]{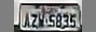} &
&   & &   
&  & \includegraphics[width=0.15\textwidth, height=0.06\textwidth]{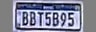} &  &  \\
&  &  & \textcolor{green}{AZW5835} &  &  &  &  &  & \textcolor{green}{BBT5B95} &  &  \\
\end{tabular}
}

\vspace{0.3mm}
\caption{Recognition results for \glspl*{lp} cropped from images with a resolution of $1280\times960$ pixels. The top row shows the predictions made by \ocrchina~\citep{liu2024irregular} on the original \gls*{lr} images, while the two subsequent rows present the predictions obtained from super-resolved images generated by \gls*{plnet}~\citep{nascimento2023super,nascimento2024superCTD} and \gls*{lcdnet}~\citep{nascimento2024enhancing}. 
Below each image, the predicted characters are displayed, with correct characters highlighted in blue and incorrect characters in red. The ground truth is indicated in green.}
\label{fig:qualitative_rsults_rr1}
\end{figure*}

    \begin{figure*}[!htb]

\vspace{0.7mm}

\centering

\resizebox{0.997\linewidth}{!}{
\begin{tabular}{
c|
c@{\hspace{0.5mm}}
c@{\hspace{0.5mm}}
c@{\hspace{0.5mm}}
c@{\hspace{0.5mm}}
c
c@{\hspace{0.5mm}}
c@{\hspace{0.5mm}}
c@{\hspace{0.5mm}}
c@{\hspace{0.5mm}}
c@{\hspace{0.5mm}}
c
}
\multirow{2}{*}{\addvbuffer[-3.5mm 8mm]{LR Inputs}} & 
\includegraphics[width=0.15\textwidth, height=0.06\textwidth]{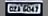} & 
\includegraphics[width=0.15\textwidth, height=0.06\textwidth]{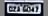} & 
\includegraphics[width=0.15\textwidth, height=0.06\textwidth]{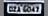} & 
\includegraphics[width=0.15\textwidth, height=0.06\textwidth]{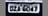} & 
\includegraphics[width=0.15\textwidth, height=0.06\textwidth]{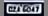} &  & 
\includegraphics[width=0.15\textwidth, height=0.06\textwidth]{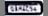} & 
\includegraphics[width=0.15\textwidth, height=0.06\textwidth]{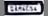} & 
\includegraphics[width=0.15\textwidth, height=0.06\textwidth]{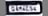} & 
\includegraphics[width=0.15\textwidth, height=0.06\textwidth]{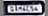} & 
\includegraphics[width=0.15\textwidth, height=0.06\textwidth]{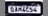} \\
& 
\textcolor{red}{OPI9C}\textcolor{blue}{47} & 
\textcolor{red}{C0A4A}\textcolor{blue}{47} & 
\textcolor{red}{OPI}\textcolor{blue}{6}\textcolor{red}{8}\textcolor{blue}{47} & 
\textcolor{red}{OZI}\textcolor{blue}{6}\textcolor{red}{C}\textcolor{blue}{47} & 
\textcolor{red}{CBI}\textcolor{blue}{6}\textcolor{red}{9}\textcolor{blue}{47} &  & 
\textcolor{blue}{Q}\textcolor{red}{I}\textcolor{blue}{M4C}\textcolor{red}{1}\textcolor{blue}{4} & 
\textcolor{red}{E}\textcolor{blue}{AM4}\textcolor{red}{E3}\textcolor{blue}{4} & 
\textcolor{red}{7}\textcolor{blue}{AM}\textcolor{red}{6}\textcolor{blue}{C}\textcolor{red}{8}\textcolor{blue}{4} & 
\textcolor{red}{PV}\textcolor{blue}{M4}\textcolor{red}{E1}\textcolor{blue}{4} & 
\textcolor{red}{OJ}\textcolor{blue}{M}\textcolor{red}{2Z5}\textcolor{blue}{4} \\
&  &  &  &  &  &  &  &  &  &  &  \\

\multirow{2}{*}{\addvbuffer[-3.5mm 8mm]{SR3}} & 
\includegraphics[width=0.15\textwidth, height=0.06\textwidth]{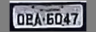} & 
\includegraphics[width=0.15\textwidth, height=0.06\textwidth]{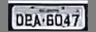} & 
\includegraphics[width=0.15\textwidth, height=0.06\textwidth]{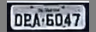} & 
\includegraphics[width=0.15\textwidth, height=0.06\textwidth]{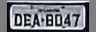} & 
\includegraphics[width=0.15\textwidth, height=0.06\textwidth]{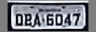} & & 

\includegraphics[width=0.15\textwidth, height=0.06\textwidth]{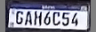} & 
\includegraphics[width=0.15\textwidth, height=0.06\textwidth]{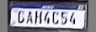} & 
\includegraphics[width=0.15\textwidth, height=0.06\textwidth]{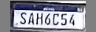} & 
\includegraphics[width=0.15\textwidth, height=0.06\textwidth]{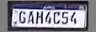} &  
\includegraphics[width=0.15\textwidth, height=0.06\textwidth]{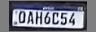} \\
& 
\textcolor{red}{OB}\textcolor{blue}{A6}\textcolor{red}{6}\textcolor{blue}{47} &     
\textcolor{blue}{D}\textcolor{red}{B}\textcolor{blue}{A6}\textcolor{red}{6}\textcolor{blue}{47} & 
\textcolor{blue}{D}\textcolor{red}{4}\textcolor{blue}{A6}\textcolor{red}{6}\textcolor{blue}{47} & 
\textcolor{blue}{D}\textcolor{red}{E}\textcolor{blue}{A}\textcolor{red}{8}\textcolor{blue}{047} & 
\textcolor{red}{QB}\textcolor{blue}{A6}\textcolor{red}{6}\textcolor{blue}{47} &  & 
 \textcolor{red}{G}\textcolor{blue}{A}\textcolor{red}{H6}\textcolor{blue}{C54} & 
 \textcolor{red}{G}\textcolor{blue}{A}\textcolor{red}{H}\textcolor{blue}{4C54} & 
 \textcolor{red}{S}\textcolor{blue}{A}\textcolor{red}{H6}\textcolor{blue}{C54} & 
 \textcolor{red}{G}\textcolor{blue}{A}\textcolor{red}{H6}\textcolor{blue}{C54} & 
 \textcolor{blue}{QA}\textcolor{red}{H6}\textcolor{blue}{C54} \\  

&  &  &  &  &  &  &  &  &  &  &  \\

\multirow{2}{*}{\addvbuffer[-3.5mm 8mm]{LPSRGAN}} & 
\includegraphics[width=0.15\textwidth, height=0.06\textwidth]{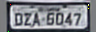} & 
\includegraphics[width=0.15\textwidth, height=0.06\textwidth]{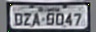} & 
\includegraphics[width=0.15\textwidth, height=0.06\textwidth]{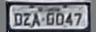} & 
\includegraphics[width=0.15\textwidth, height=0.06\textwidth]{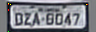} & 
\includegraphics[width=0.15\textwidth, height=0.06\textwidth]{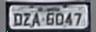} & & 

\includegraphics[width=0.15\textwidth, height=0.06\textwidth]{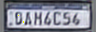} & 
\includegraphics[width=0.15\textwidth, height=0.06\textwidth]{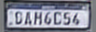} & 
\includegraphics[width=0.15\textwidth, height=0.06\textwidth]{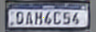} & 
\includegraphics[width=0.15\textwidth, height=0.06\textwidth]{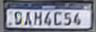} &  
\includegraphics[width=0.15\textwidth, height=0.06\textwidth]{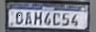} \\
& 
\textcolor{blue}{DZA6047} & 
\textcolor{blue}{DZA6047} & 
\textcolor{blue}{DZA6047} & 
\textcolor{red}{O}\textcolor{blue}{ZA}\textcolor{red}{8}\textcolor{blue}{047} & 
\textcolor{blue}{DZA6047} &  & 

\textcolor{red}{O}\textcolor{blue}{AM}\textcolor{red}{6}\textcolor{blue}{C54} & 
\textcolor{red}{O}\textcolor{blue}{AM}\textcolor{red}{6}\textcolor{blue}{C54} & 
\textcolor{red}{O}\textcolor{blue}{A}\textcolor{red}{H6}\textcolor{blue}{C54} & 
\textcolor{blue}{QAM4C54} & 
\textcolor{red}{O}\textcolor{blue}{AM}\textcolor{red}{6}\textcolor{blue}{C54} \\ 

&  &  &  &  &  &  &  &  &  &  &  \\

\multirow{2}{*}{\addvbuffer[-3.5mm 8mm]{Real-ESRGAN}} & 
\includegraphics[width=0.15\textwidth, height=0.06\textwidth]{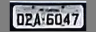} & 
\includegraphics[width=0.15\textwidth, height=0.06\textwidth]{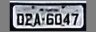} & 
\includegraphics[width=0.15\textwidth, height=0.06\textwidth]{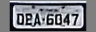} & 
\includegraphics[width=0.15\textwidth, height=0.06\textwidth]{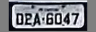} & 
\includegraphics[width=0.15\textwidth, height=0.06\textwidth]{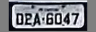} & & 

\includegraphics[width=0.15\textwidth, height=0.06\textwidth]{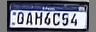} & 
\includegraphics[width=0.15\textwidth, height=0.06\textwidth]{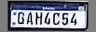} & 
\includegraphics[width=0.15\textwidth, height=0.06\textwidth]{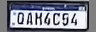} & 
\includegraphics[width=0.15\textwidth, height=0.06\textwidth]{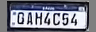} &  
\includegraphics[width=0.15\textwidth, height=0.06\textwidth]{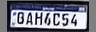} \\
& 
\textcolor{blue}{DZA6}\textcolor{red}{6}\textcolor{blue}{47} & 
\textcolor{blue}{DZA6}\textcolor{red}{6}\textcolor{blue}{47} & 
\textcolor{blue}{DBA6}\textcolor{red}{6}\textcolor{blue}{47} & 
\textcolor{blue}{DZA6}\textcolor{red}{6}\textcolor{blue}{47} & 
\textcolor{red}{O}\textcolor{blue}{ZA6}\textcolor{red}{6}\textcolor{blue}{47} &  & 
\textcolor{blue}{QA}\textcolor{red}{H6}\textcolor{blue}{C54} & 
\textcolor{blue}{QA}\textcolor{red}{H6}\textcolor{blue}{C54} & 
\textcolor{blue}{QA}\textcolor{red}{M}\textcolor{blue}{4C54} & 
\textcolor{red}{O}\textcolor{blue}{AH4C54} & 
\textcolor{red}{O}\textcolor{blue}{AH}\textcolor{red}{6}\textcolor{blue}{C54} \\  

&  &  &  &  &  &  &  &  &  &  &  \\

\multirow{2}{*}{\addvbuffer[-3.5mm 8mm]{\gls*{plnet}}} & 
\includegraphics[width=0.15\textwidth, height=0.06\textwidth]{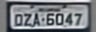} & 
\includegraphics[width=0.15\textwidth, height=0.06\textwidth]{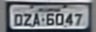} & 
\includegraphics[width=0.15\textwidth, height=0.06\textwidth]{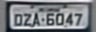} & 
\includegraphics[width=0.15\textwidth, height=0.06\textwidth]{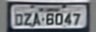} &  
\includegraphics[width=0.15\textwidth, height=0.06\textwidth]{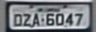} &  & 

\includegraphics[width=0.15\textwidth, height=0.06\textwidth]{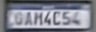} & 
\includegraphics[width=0.15\textwidth, height=0.06\textwidth]{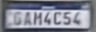} & 
\includegraphics[width=0.15\textwidth, height=0.06\textwidth]{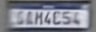} & 
\includegraphics[width=0.15\textwidth, height=0.06\textwidth]{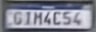} &  
\includegraphics[width=0.15\textwidth, height=0.06\textwidth]{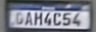} \\
& 
\textcolor{red}{O}\textcolor{blue}{ZA}\textcolor{blue}{6047} & 
\textcolor{red}{O}\textcolor{blue}{ZA}\textcolor{blue}{6047} & 
\textcolor{red}{O}\textcolor{blue}{ZA}\textcolor{blue}{6047} & 
\textcolor{red}{O}\textcolor{blue}{ZA}\textcolor{red}{8}\textcolor{blue}{047} & 
\textcolor{red}{O}\textcolor{blue}{ZA6047}  &  & 

\textcolor{red}{D}\textcolor{blue}{AM4C54} & 
\textcolor{red}{G}\textcolor{blue}{AM4C54} & 
\textcolor{red}{S}\textcolor{blue}{AM4C54} & 
\textcolor{red}{GI}\textcolor{blue}{M4C54} & 
\textcolor{blue}{QA}\textcolor{red}{H}\textcolor{blue}{4C54}  \\
&  &  &  &  &  &  &  &  &  &  &  \\
\multirow{2}{*}{\addvbuffer[-3.5mm 8mm]{\gls*{lcdnet}}} & 
\includegraphics[width=0.15\textwidth, height=0.06\textwidth]{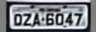} & 
\includegraphics[width=0.15\textwidth, height=0.06\textwidth]{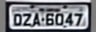} & 
\includegraphics[width=0.15\textwidth, height=0.06\textwidth]{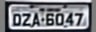} & 
\includegraphics[width=0.15\textwidth, height=0.06\textwidth]{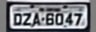} & 
\includegraphics[width=0.15\textwidth, height=0.06\textwidth]{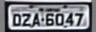} & & 

\includegraphics[width=0.15\textwidth, height=0.06\textwidth]{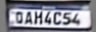} & 
\includegraphics[width=0.15\textwidth, height=0.06\textwidth]{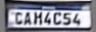} & 
\includegraphics[width=0.15\textwidth, height=0.06\textwidth]{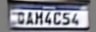} & 
\includegraphics[width=0.15\textwidth, height=0.06\textwidth]{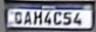} &  
\includegraphics[width=0.15\textwidth, height=0.06\textwidth]{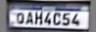} \\
& 
\textcolor{red}{Q}\textcolor{blue}{ZA6047} & 
\textcolor{red}{O}\textcolor{blue}{ZA6047}& 
\textcolor{red}{O}\textcolor{blue}{ZA6047} & 
\textcolor{red}{Q}\textcolor{blue}{ZA}\textcolor{red}{8}\textcolor{blue}{047} & 
\textcolor{red}{O}\textcolor{blue}{ZA6047} &  & 
\textcolor{blue}{QAM4C54} & 
\textcolor{red}{C}\textcolor{blue}{AM4C54} & 
\textcolor{blue}{QAM4C54} & 
\textcolor{blue}{QA}\textcolor{red}{H}\textcolor{blue}{4C54} & 
\textcolor{blue}{QA}\textcolor{red}{H}\textcolor{blue}{4C54} \\  

\hdashline

&  &  &  &  &  &  &  &  &  &  &  \\

\multirow{2}{*}{\addvbuffer[-3.5mm 8mm]{HR}} &   
&  & \includegraphics[width=0.15\textwidth, height=0.06\textwidth]{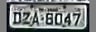} &
&   & &   
&  & \includegraphics[width=0.15\textwidth, height=0.06\textwidth]{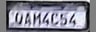} &  &  \\
&  &  & \textcolor{green}{DZA6047} &  &  &  &  &  & \textcolor{green}{QAM4C54} &  &  \\
\end{tabular}
}

\vspace{0.3mm}
\caption{Recognition results for \glspl*{lp} cropped from images with a resolution of $1920\times1080$ pixels. The top row shows the predictions made by \ocrchina~\citep{liu2024irregular} on the original \gls*{lr} images, while the two subsequent rows present the predictions obtained from super-resolved images generated by \gls*{plnet}~\citep{nascimento2023super,nascimento2024superCTD} and \gls*{lcdnet}~\citep{nascimento2024enhancing}. 
Below each image, the predicted characters are displayed, with correct characters highlighted in blue and incorrect characters in red. The ground truth is indicated in green.}
\label{fig:qualitative_rsults_rr2}
\end{figure*}

\fi

\cref{fig:qualitative_rsults_rr1,fig:qualitative_rsults_rr2} present comparisons between low-resolution \gls*{lp} images and their super-resolved counterparts produced by the super-resolution networks. 
As expected, the \gls*{gplpr} model performs worse on low-resolution \glspl*{lp} (i.e., before super-resolution) extracted from $1280 \times 960$ images~(\cref{fig:qualitative_rsults_rr1}) compared to $1920 \times 1080$ images~(\cref{fig:qualitative_rsults_rr2}).
This performance gap stems primarily from the reduced pixel density in the smaller images, which causes characters to blend into the \gls*{lp} background.
This effect is evident in the super-resolved \glspl*{lp} shown in \cref{fig:qualitative_rsults_rr1}, where a ``W'' is mistakenly reconstructed as a ``K'' in the Brazilian \gls*{lp}~(left), and a ``B'' is reconstructed as ``R'' or ``H'' in the Mercosur \gls*{lp}~(right).
Similar errors are observed in \cref{fig:qualitative_rsults_rr2},  though less frequently due to the higher pixel density.
For example, a ``Q'' was reconstructed as an ``O,'' and an ``M'' was reconstructed as a ``H,'' with \gls*{plnet}-generated images exhibiting a more pronounced tendency towards these misclassifications than \gls*{lcdnet}-generated~images.

\subsubsection{Runtime Analysis}

\major{This section evaluates the inference efficiency and suitability of super-resolution models for both real-time and forensic applications.
\Cref{tab:inference_clean} presents a quantitative comparison of inference times and computational performance across the models.
To ensure consistency and reliability, each model was independently tested over five separate~runs.}

\begin{table}[!htb]
\centering
\caption{Inference Time and FPS Comparison.}
\label{tab:inference_clean}
\resizebox{0.85\linewidth}{!}{
\begin{tabular}{@{}lccc@{}}
\toprule
\multirow{2}{*}{\textbf{Model}}  & \textbf{Time (ms)} & \textbf{Time (ms)} & \textbf{FPS} \\
 & \textbf{Avg ± Std} & \textbf{Min - Max} & \textbf{Avg} \\ 
\midrule
\acrshort*{lpsrgan}  & \phantom{0}4.82 ± 0.45  & \phantom{0}4.49 - \phantom{0}7.41 & 207.5\\

\acrshort*{plnet}  & 24.88 ± 1.21 & 23.97 - 31.68 & \phantom{0}40.2 \\

\acrshort*{realesrgan}  & 34.86 ± 2.66  & 32.53 - 46.46 & \phantom{0}28.7 \\

\gls*{lcdnet}  & 61.88 ± 1.77 & 60.47 - 73.13 & \phantom{0}16.1 \\
\bottomrule

\end{tabular}
}
\end{table}

\major{\Acrshort*{lpsrgan} emerges as the fastest model, processing \(207.5\)~\gls*{fps} (\(4.82 \pm 0.45\)~\si{\milli\second}), making it well-suited for real-time applications such as traffic monitoring.
However, this speed comes at the expense of performance: as shown in \cref{xp:quantitativeR35MJ}, the \ocrchina model achieves only a $28.8$\% recognition rate when applied to images super-resolved by \acrshort*{lpsrgan}, considerably lower than the rates obtained with slower \gls*{sr} models.
In contrast, \gls*{lcdnet} leads to the highest recognition rate (\(42.3\%\) with \gls*{mvcp} fusion on 5 images), a \(1.4\times\) improvement over \acrshort*{lpsrgan}, despite its slower inference speed (\(61.88\)~\si{\milli\second}, \(16.1\)~\gls*{fps}).  
\Gls*{plnet} offers a balanced trade-off, combining moderate efficiency (\(24.88~\pm~1.21\)~\si{\milli\second}, \(40.2\)~\gls*{fps}) with competitive performance (\(40.9\%\) recognition rate using \gls*{mvcp}), while \acrshort*{realesrgan} performs poorly in both inference speed (\(34.86\)~\si{\milli\second}, \(28.7\)~\gls*{fps}) and recognition performance (\(29.5\%\) recognition rate with \gls*{mvcp}).}

\major{Notably, \gls*{lcdnet} exhibits stable inference times (standard deviation:~\(1.77\)~\si{\milli\second}), contrasting with \acrshort*{lpsrgan}'s higher variability (ranging from~\(4.49\)~to~\(7.41\)~\si{\milli\second}).  
This stability, coupled with its superior accuracy under the \gls*{mvcp} protocol, establishes \gls*{lcdnet} as a strong candidate for forensic applications, where precision significantly outweighs speed requirements. 
These findings highlight a key trade-off: while latency-sensitive deployments may favor lightweight models such as \acrshort*{lpsrgan}, accuracy-critical scenarios benefit more from specialized architectures like \gls*{lcdnet}, even at the cost of increased computational~demands.}

\major{We also evaluated the SR3 model,  which showed considerably higher inference times (\(
12.31 \pm 0.018~\si{\second}\)).
Although impractical for real-time use, SR3 may still be viable in forensic contexts where processing time is less restrictive.
Due to its significantly slower performance --~operating in seconds rather than milliseconds~-- SR3's results are omitted from \cref{tab:inference_clean}, as they fall outside the real-time operational scope targeted by the other models.}

\section{Conclusions and Future Directions}
\label{sec:Conclusions}

\glsreset{lpr}
\glsreset{mvcp}

In this work, we introduced \srplates, a publicly available dataset specifically designed for \gls*{lp} super-resolution.
The dataset comprises $100{,}000$ \gls*{lp} images organized into $10{,}000$ tracks, each containing five sequential \gls*{lr} and five sequential \gls*{hr} images of the same \gls*{lp}.
This dataset is highly valuable for developing and evaluating super-resolution techniques aimed at improving \gls*{lpr} under real-world, low-quality conditions.
Although primarily intended for \gls*{lp} super-resolution, \dataset is also well-suited for training and evaluating \gls*{lpr} models, as it currently represents the largest collection of Brazilian and Mercosur \glspl*{lp}~available.

\major{We proposed a benchmark using the \dataset dataset.
We assessed the recognition rates achieved by the \ocrchina model~\citep{liu2024irregular} in conjunction with five super-resolution approaches: general-purpose models (\acrshort*{sr3}~\citep{saharia2023image} and \acrshort*{realesrgan}~\citep{wang2021real}) and \gls*{lp}-specialized networks (\acrshort*{lpsrgan}, \gls*{plnet}~\citep{nascimento2023super}, and \gls*{lcdnet}~\citep{nascimento2024enhancing}). By fusing predictions from multiple super-resolved images via \gls*{mvcp}, recognition rates improved from $2.2\%$ (raw \gls*{lr} images) to $29.9\%$ for single-image outputs and up to $42.3\%$ with five-image fusion using \gls*{lcdnet}.
This represents a $19.2\times$ improvement over raw \gls*{lr} images, with \gls*{mvcp} outperforming other fusion strategies by $3.8–8.1\%$, demonstrating its robustness in aggregating temporal information. Notably, \gls*{lcdnet} consistently surpassed both general-purpose and \gls*{lp}-specialized alternatives, validating the importance of architectural designs tailored to character structure and layout preservation.}

Our findings offer valuable insights into the benefits of super-resolving multiple low-resolution versions of the same \gls*{lp} and combining their \gls*{ocr} predictions.
This method is especially beneficial for surveillance and forensic applications, where partial matches can greatly reduce the search space for potential \glspl*{lp}.
We believe this approach could be further strengthened by incorporating additional vehicle attributes, such as make, model, and color, as suggested by previous research~\citep{oliveira2021vehicle,lima2024toward}.

\major{For future work, we aim to enhance \gls*{lp} domain-specific architectures by incorporating multi-image fusion to enable temporal learning.
Additionally, we plan to expand the \dataset dataset to address its current limitations: 
(i)~Motorcycle \glspl{lp}, which were excluded due to their two-row layout and acquisition challenges --~such as the relatively low traffic of motorcycles at our capture site and frequent repetitions of the same \glspl*{lp};
(ii)~Systematic \gls*{ocr} errors (e.g., confusion between ``B'' and~``8''), which we intend to mitigate using multi-model ensembling or confidence calibration techniques;  
(iii) Regional and nighttime \glspl*{lp}, as the current dataset is limited to daylight Brazilian/Mercosur \glspl*{lp} --~we aim to align it with recent multi-region \gls*{alpr} systems~\citep{laroca2022cross};
(iv) Generalization to extreme conditions (e.g., rain, haze) via hybrid synthetic-real training.}

\major{We remark that these limitations do not undermine \dataset's value as a foundational benchmark for real-world \gls*{sr} research.
On the contrary, these limitations underscore clear opportunities for advancing robust \gls*{alpr} systems.
By publicly releasing \dataset, we aim to drive progress in super-resolution and recognition of degraded \glspl*{lp}, particularly in unconstrained scenarios where existing synthetic benchmarks prove~insufficient.
}

\section*{Acknowledgments}

This study was financed in part by the \textit{Coordenação de Aperfeiçoamento de Pessoal de Nível Superior - Brasil~(CAPES)} - Finance Code 001, and in part by the \textit{Conselho Nacional de Desenvolvimento Científico e Tecnológico~(CNPq)} (\#~315409/2023-1 and \#~312565/2023-2).
We gratefully acknowledge the support of NVIDIA Corporation with the donation of the Quadro RTX $8000$ GPU used for this research.

\section*{Declarations}

\begin{contributions}

Valfride Nascimento is the main contributor and writer of this manuscript.
Gabriel E. Lima contributed to experiment validation and manuscript review.
Rafael O. Ribeiro provided project resources and contributed to the manuscript review process.
William R. Schwartz co-supervised the project and reviewed the manuscript.
Rayson Laroca contributed to the conceptualization and methodology and assisted in the review and editing of the manuscript.
David Menotti supervised the project, provided project resources, and contributed to the review and editing of the manuscript.
All authors read and approved the final~manuscript.

\end{contributions}

\balance

\begin{interests}
The authors declare that they have no competing interests.
\end{interests}

\begin{materials}
The resources generated and analyzed during this study are available at~\textbf{\url{https://valfride.github.io/nascimento2024toward/}}.

\end{materials}

\bibliographystyle{apalike-sol}
\bibliography{bibtex}

\end{document}